\documentclass[10pt,letter,journal]{IEEEconf}
\usepackage{amsmath,amsfonts,amssymb}
\let\labelindent\relax
\usepackage{prettyref}
\usepackage{mathrsfs}
\usepackage{graphicx}
\usepackage{wrapfig}
\usepackage{MnSymbol}
\usepackage{multirow}
\usepackage{booktabs}
\usepackage{algorithm}
\usepackage{enumitem}
\usepackage[noend]{algpseudocode}
\usepackage[table]{xcolor}
\usepackage{cellspace}
\usepackage{mathtools}
\usepackage{sidecap}
\usepackage[font=small]{caption}
\usepackage{subcaption}
\usepackage{hyperref}
\usepackage
[backend=bibtex,
bibstyle=ieee,
citestyle=numeric,
mincitenames=1,
maxcitenames=2,
natbib=true,
doi=false,
isbn=false,
url=false,
eprint=false]{biblatex}

\newcommand{\citewithauthor}[1]{\citeauthor{#1} \cite{#1}}

\algnewcommand{\LineComment}[1]{\State \(\triangleright\) #1}
\algdef{SE}[DOWHILE]{Do}{doWhile}{\algorithmicdo}[1]{\algorithmicwhile\ #1}
\newcommand*{\colorboxed}{}
\def\colorboxed#1#{%
  \colorboxedAux{#1}%
}
\newcommand*{\colorboxedAux}[3]{%
  \begingroup
    \colorlet{cb@saved}{.}%
    \color#1{#2}%
    \boxed{%
      \color{cb@saved}%
      #3%
    }%
  \endgroup
}

\newrefformat{Fig}{Figure~\ref{#1}}
\newrefformat{fig}{Figure~\ref{#1}}
\newrefformat{par}{Section~\ref{#1}}
\newrefformat{appen}{Appendix~\ref{#1}}
\newrefformat{ass}{Assumption~\ref{#1}}
\newrefformat{sec}{Section~\ref{#1}}
\newrefformat{sub}{Section~\ref{#1}}
\newrefformat{table}{Table~\ref{#1}}
\newrefformat{alg}{Algorithm~\ref{#1}}
\newrefformat{Alg}{Algorithm~\ref{#1}}
\newrefformat{Def}{Definition~\ref{#1}}
\newrefformat{Thm}{Theorem~\ref{#1}}
\newrefformat{Lem}{Lemma~\ref{#1}}
\newrefformat{step}{Step~\ref{#1}}
\newrefformat{ln}{Line~\ref{#1}}
\newrefformat{eq}{Equation~\ref{#1}}
\newrefformat{pb}{Problem~\ref{#1}}
\newrefformat{it}{Item~\ref{#1}}
\newrefformat{te}{Term~\ref{#1}}
\def\Eqref Eq:#1:{\eqref{eq:#1}}
\newrefformat{Eq}{Equation~\Eqref#1:}

\newcommand{\E}[1]{\mathbf{#1}}
\newcommand{\TE}[1]{\textbf{#1}}

\newcommand{\argmin}[1]{\underset{#1}{\E{argmin}}\;}

\newcommand{\argminP}[1]{\E{argmin}\;}

\newcommand{\argmaxP}[1]{\E{argmax}\;}
\newcommand{\ST}{\E{s.t.}\;}

\definecolor{darkgreen}{HTML}{186a3b}

\newif\ifarxiv
\arxivtrue
\def\BibTeX{{\rm B\kern-.05em{\sc i\kern-.025em b}\kern-.08em
    T\kern-.1667em\lower.7ex\hbox{E}\kern-.125emX}}
\begin{document}

\title{\LARGE \bf Optimized Coverage Planning for UV Surface Disinfection}
\author{Jo\~ao Marcos Correia Marques$^{1}$, Ramya Ramalingam$^{2}$, Zherong Pan$^{1}$ and Kris Hauser$^{1}$ \thanks{*J. Marques and K. Hauser are partially supported by NSF Grant NRI-2025782. R. Ramalingam is supported by a CRA-WP DREU grant. $^{1}$J. M. C. Marques, Z. Pan and K. Hauser are with the Department of Computer Science, University of Illinois at Urbana-Champaign, Urbana, IL, USA {\tt\small \{jmc12,zherong,kkhauser\}@illinois.edu}. $^{2}$R. Ramalingam is with the Dept. of Computer Science and the Dept. of Mathematics, Harvey Mudd College, Claremont, CA, USA {\tt\small rramalingam@g.hmc.edu}.}}

\maketitle

\begin{abstract}
UV radiation has been used as a disinfection strategy to deactivate a wide range of pathogens, but existing irradiation strategies do not ensure sufficient exposure of all environmental surfaces and/or require long disinfection times. We present a near-optimal coverage planner for mobile UV disinfection robots. The formulation optimizes the irradiation time efficiency, while ensuring that a sufficient dosage of radiation is received by each surface. The trajectory and dosage plan are optimized taking collision and light occlusion constraints into account. We propose a two-stage scheme to approximate the solution of the induced NP-hard optimization, and, for efficiency, perform key irradiance and occlusion calculations on a GPU. Empirical results show that our technique achieves more coverage for the same exposure time as strategies for existing UV robots, can be used to compare UV robot designs, and produces near-optimal plans. This is an extended version of the paper originally contributed to ICRA2021.
\end{abstract}


\section{Introduction}
The Covid-19 pandemic has encouraged worldwide innovation in methods for reducing the risk of disease transmission in hospitals, public transportation and other public spaces. One promising technology is ultraviolet (UV) disinfection of surfaces, which has strong antimicrobial properties particularly in the UVC (200--280\,nm) spectrum. UVC has long been known to deactivate a wide range of pathogens, such as Coronaviruses \cite{bedell2016efficacy,Heling2020UltravioletStudies.}, bacteria and protozoans \cite{Hijnen2006InactivationReviewb}. 
Existing UV delivery approaches include air and water disinfection systems used in filtration and waste processing plants \cite{Heling2020UltravioletStudies.}, as well as surface disinfection systems in the form of wands \cite{lyon2008uv}, overhead lights, pushcarts, and mobile robots carrying high-power UVC lamps \cite{park2007robot}. Hospital testing~\cite{armellino2020comparative} has shown that a combination of standard manual cleaning followed by UVC surface irradiation has shown to be more effective in disinfecting environments than manual cleaning alone. 

\begin{figure}[t!] 
\centering
\includegraphics[width=0.8\linewidth]{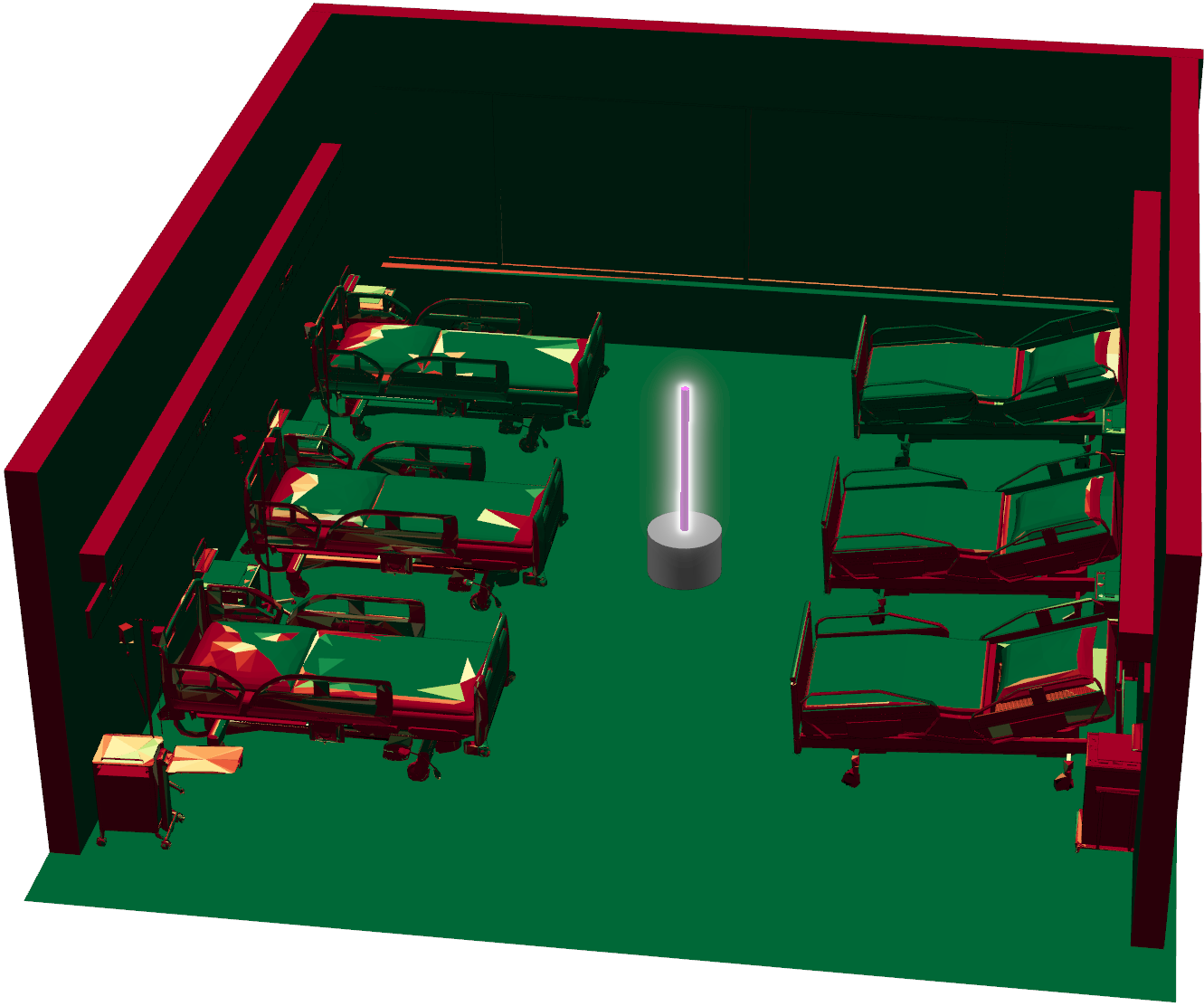}
\includegraphics[width=0.8\linewidth]{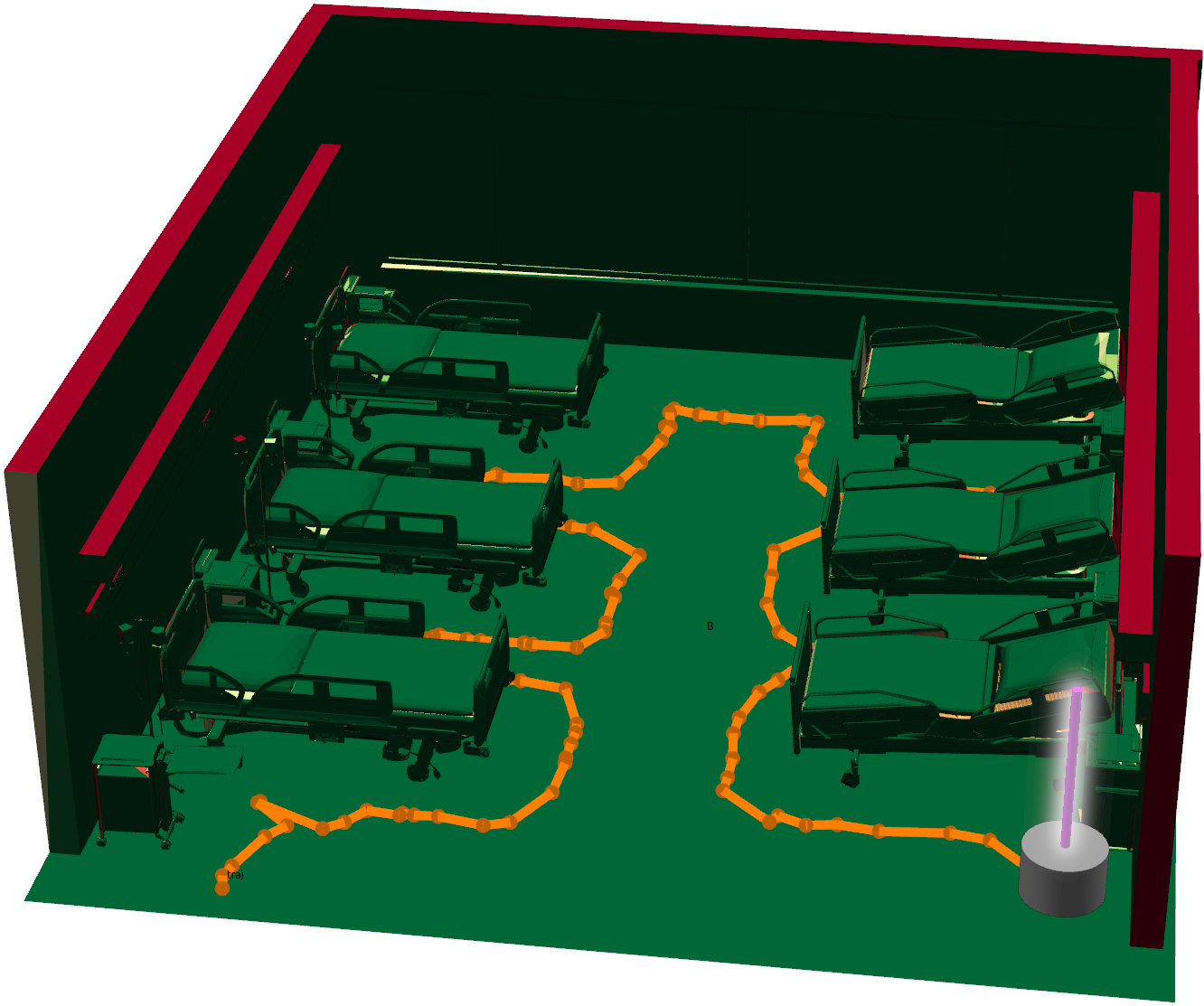}
\caption{\label{fig: single point 3D disnfection} Comparison of a standard stationary mobile robot (top) against an optimized motion (bottom). Robot carries a tower light that emits UV radiation uniformly along its length. Surfaces are color coded by UV fluence received, with red indicating 0\,J/m$^2$ and green indicating 280\,J/m$^2$ or higher. A stationary light is unable to disinfect much of the environment after $30$ minutes, while a mobile robot following our optimally computed trajectory (in orange) achieves almost complete coverage. (Best viewed in color)}
\end{figure}

Dosing is an important factor in effective use of UVC, and is usually performed by following manufacturers' guidelines. Although some UV disinfection robots also feature sensors that measure reflected radiant energy as an approximation of surface dosage, existing methods fail to disinfect certain parts of the environment~\cite{Lindblad2019Ultraviolet-CNeeded}.
Two pitfalls are noted. The radiant fluence received by a surface is affected by the inverse square law, so fluence drops quickly as distance increases. 
Second, occlusions also affect the delivery of light into back-facing or shadowed regions. 
These effects are illustrated in \prettyref{fig: single point 3D disnfection}, which shows a simulation of the irradiation of a hospital infirmary under a static UV tower, demonstrating ineffective disinfection of bedsides and occluded equipment.

We present a method for planning optimal trajectories of a mobile UV disinfection robot with dosing constraints. Our optimization can be configured to prioritize coverage of high-touch surfaces under a fixed time budget, or to guarantee the eventual full disinfection of all surfaces reachable by irradiation. The robot's movement must be collision-free while conforming to the dosing constraints. We solve the problem by building a probabilistic roadmap in the robot's configuration space, and then finding a tour of a subset of configurations that optimizes the dose. The coverage problem on the roadmap can be cast as an NP-hard Mixed-Integer Linear Programming (MILP), but we propose an approximate two-stage solver that uses a Linear Program (LP) to find dwell times followed by a Traveling Salesman Problem (TSP) to find the tour. Experiments show that our solver is orders of magnitude faster than MILP with a loss of less than $3\%$ of optimality.  Moreover, dosage planning requires determination of an {\em irradiance matrix} that considers visibility and exposure of every surface patch from each candidate UV light pose, and we propose an approach that efficiently calculates this large matrix using a Graphics Processing Unit (GPU). 
\section{\label{sec:related}Related Work}
Motion planning for UV disinfection bears a resemblance to two well-studied problems: coverage and inspection planning. The goal of coverage planning \cite{Cheng2008Time-optimalCoverage,das_probably_approximately_2011,Paull2013Sensor-drivenVehicles,galceran2015coverage} is for every point in the freespace to be covered by the robot, while the goal of inspection planning \cite{Englot2010InspectionStructures,Heng2015EfficientEnvironments,Bircher2015StructuralRobotics,bircher2017incremental,bogaerts2018gradient} is for every point on an object surface to be visible from some point on the robot trajectory. The disinfection planning problem introduced in this paper adds an additional layer of complexity to inspection planning, where every point in an object surface must receive a certain amount of irradiance exposure. This scenario induces a joint problem of robot trajectory planning and disinfection time assignment. Compared with standard coverage and inspection planning, UV disinfection is applied routinely in healthcare facilities, public spaces, and food industries, and can take tens of minutes to ensure enough dosage. Therefore, achieving (near) optimality in reducing the disinfection time for a known environment, which is the focus of this paper, is more important than adapting to unknown environments or online re-planning as done in Refs.~\cite{Heng2015EfficientEnvironments,galceran2015coverage}.

Besides robotics, UV disinfection planning can be understood as an effort to model and control light transport. In this aspect, there is overlap with similar efforts in the field of radiation dosage planning \cite{lee2003integer,bahr1968method,ezzell1996genetic,romeijn2006new}, rendering of Lambertian surfaces using boundary element method \cite{sillion1994radiosity,keller1997instant,coombe2004radiosity,laine2007incremental} (otherwise known as radiosity), and optimization of light placements \cite{stefan2013analysis,zhang2013lighting}. The radiation dosage planning problem has the same goal as our problem, ensuring the delivery of sufficient amount of dosage to target volumes. An additional goal is to reduce the dosage as much as possible for the organs at risk. However, since geometric information of human organs is difficult to acquire, these methods are mostly heuristic and sub-optimal. Radiosity is used to only model light transportation, reflection, and absorption. Of particular interest is GPU-accelerated radiosity \cite{coombe2004radiosity} where the occlusion map is computed using GPU rasterization. A similar technique is used in this work, while indirect light reflections are ignored by our method as their contributions are assumed neglectable. Other works on lighting optimization for urban design or scientific data visualization \cite{stefan2013analysis,zhang2013lighting} also considers moving light sources, but these lights are fixed after the design phase.

Finally, \citewithauthor{Hu2020SegmentingEnvironments} present a pipeline for UV disinfection of built environments, wherein 3D environments are reconstructed using off-the-shelf SLAM, while performing affordance estimation using a neural-network on the RGBD streams and projecting them into the generated map. The labels are then projected onto the 3D map and used to indicate areas that are likely to be contaminated and direct their robot to those segments, executing one of two disinfection motion primitives on the environment - a scanning motion for flat segments and a circling motion for cylindrical shapes. They do not, however, provide any reasoning or experiments regarding the effectiveness of their disinfection strategy, focusing their results on the performance of the affordance estimates and on the mechanical feasibility of their disinfection primitives on a static robot. Their planning strategy also does not consider a disinfection tour, but limits itself to planning collision-free trajectories between the robot's current position and a given area to be disinfected. 
\section{\label{sec:model}UV Disinfection Trajectory Planning}
Here we formalize the path planning problem for targeted UV disinfection first as a continuous, infinite-dimensional trajectory optimization problem, and then as a discrete approximation. 

\subsection{\label{sec:continuous}Continuous Formulation}
Let $\mathbb{E} \subset \mathbb{R}^3$ be the boundary of the environment, which is the surface to be disinfected. The disinfection is performed using a mobile robot equipped with a UV light, where $\mathbb{C}$ is the robot's configuration space and $\mathbb{C}_{free}$ is the freespace. When the robot assumes any collision-free configuration $x\in\E{C}_{free}$, each infinitesimal surface patch $ds \in \mathbb{E}$ will receive a certain amount of radiative fluence per second. We model the radiative fluence distribution using a so-called Poynting vector function $\mathbb{I}(x,ds)$, such that the infinitesimal surface patch $ds$ receives the following irradiance:
\begin{equation}
\label{eq:Continuous Irradiance}
I_{ds}(x) = \langle\mathbb{I}(x,ds),n(s)\rangle,
\end{equation}
where $ds$ is the infinitesimal surface patch with outward normal $n(s)$ and $\langle\bullet,\bullet\rangle$ is the inner product. Note that $\mathbb{I}(x,ds)$ already encodes the effects of light mirror reflections and occlusions by the environment. For instance, in the case where there are full occlusions before reaching $ds$, this vector is zero. We denote $\tau(t): \mathbb{R} \mapsto \mathbb{C}_{free}$ as the trajectory in the robot configuration space parameterized in time $t\in [0,T_{final}]$. The radiant fluence (also known as radiant exposure) of an infinitesimal surface patch $ds$ from a trajectory $\tau$, denoted by $\mu_{ds}$, is described by:
\begin{equation}
\label{eq:continous flunce equation}
\mu_{ds}(\tau) = \int_{0}^{T_{final}}I_{ds}(\tau(t))dt.
\end{equation}
We define the minimum-time, continuous path planning problem for UV disinfection as:
\begin{equation}
\begin{aligned}
\label{eq:continuous minimization problem}
\argmin{T_{final},\tau}&T_{final}   \\
\ST\;&\mu_{ds}(\tau) \geq \mu_{min}(ds) \quad \forall ds \\
&\forall t\in[0,T_{final}]
\begin{cases}
\tau(t)\in\mathbb{C}_{free}&   \\
\dot{\tau}(t)=f(\tau,\dot{\tau},u)&  \\
\|u(t)\|\leq u_{max}&    \\
\end{cases},
\end{aligned}
\end{equation}
where $f(\tau,\dot{\tau},u)$ encodes the robot dynamics , $u(t)$ is the control signal, $u_{max}$ is the control limits and $\mu_{min}(ds)$ is the minimum disinfection fluence (dose) prescribed to the surface. The prescribed dose can be surface-dependent (e.g., to deliver more radiation to high-touch surfaces), but we set a constant $\mu_{min}$ for notational simplicity. Eq.~\ref{eq:continuous minimization problem} is intractable due to the infinite number of constraints and the integral in \prettyref{eq:continous flunce equation}.

\subsection{\label{sec:discrete}Discrete Formulation}
Next, we formulate a discrete counterpart of \eqref{eq:continuous minimization problem}. The surface $\mathbb{E}$ is discretized using a simplicial complex with $N$ triangles, $\{s_i|i=1,\cdots,N\}$. The robot can only take a discrete set of $K$ configurations $\{x_1,\cdots,x_K\}\subset\mathbb{C}_{free}$. Each configuration $x_k$ is called a vantage configuration. To simplify total irradiance calculations, we assume that the light source stops at each configuration $x_k$ in its trajectory for some dwelling time, denoted as $t_k\geq0$, and emits no radiation during the transition between vantage configurations. Let $\mathbf{t}$ be the vector of $K$ dwell times. We then discretize~\eqref{eq:Continuous Irradiance} and  \eqref{eq:continous flunce equation}  as:
\begin{equation}
\label{eq:Discrete Irradiance}
I_i(x_k)=\int_{s_i}\langle\mathbb{I}(x_k,ds),n(s)\rangle ds,
\end{equation}
\begin{equation}
\label{eq:Discrete flunce equation}
\mu_i(\mathbf{t})=\sum_{k=1}^KI_i(x_k)t_k.
\end{equation}
Suppose there exists a network of paths between configurations that satisfies kinematics and dynamics constraints. Let $d_{kl}\geq0$ be the distance along the network between any $x_k$ and $x_l$, with $d_{kl}=\infty$ if no path connects them. We then formulate the discrete version of \eqref{eq:continuous minimization problem} as a path subset selection problem. We introduce binary variables $z_{kl}\in\{0,1\}$, each indicating whether the path $d_{kl}$ is used in the final path, and a vector $\mathbf{z}$ collecting each indicator. Then the discrete version of \eqref{eq:continuous minimization problem} is defined as:
\begin{equation}
\begin{aligned}
\label{eq:discrete general trajectory minimization}
\argmin{\mathbf{t},\mathbf{z}}&\sum_{k=1}^Kt_k + \frac{1}{v_{max}}\sum_{k=1}^K \sum_{l=1}^K d_{kl}z_{kl} \\
\ST&\mu_i(\mathbf{t}) \geq \mu_{min}\quad\forall i=1,\cdots,N \\ 
&\mathbf{z}\text{ connected}  \\
&t_k>0\text{ iff $z_{kl}=1$ or $z_{lk}=1$ for some $l$}. 
\end{aligned}
\end{equation}
The last two conditions are consistency constraints, stating that the selected paths form a simply connected path, and the second ensures that the robot can only dwell on vantage configurations that are part of the selected path. 

\section{Proposed Solution}

In this section, we propose a novel approximate algorithm to search for near-optimal coverage plans.  The main steps of our approach are listed below:
\begin{enumerate}
\item Select vantage points $\{x_1,\cdots,x_{K'}\}$ and obtain the subset of {\em vantage configurations} $\{q_1,\cdots,q_K\}$ for feasible points (Detailed in Section IV-A).
\item Compute network $\mathcal{R}$ of paths between configurations using a PRM-style approach. Retain subset of reachable configurations. (Sec.IV-B)
\item Compute irradiance matrix $I_i(x_k)$  (Sec.IV-C)
\item Solve a LP for optimal dwell times $\mathbf{t}$ (Sec.IV-D)
\item Solve a TSP for a tour of all configurations $q_k$ for which dwell time is nonzero, that is $t_k>0$ (Sec.IV-D)
\item Execute the tour, stopping for time $t_k$ at each visited configuration $q_k$
\end{enumerate}

We show in \ifarxiv Appendix I\else extended report \cite{extended_report}\fi, that Eq.~\eqref{eq:discrete general trajectory minimization} can be formulated as a Mixed Integer Linear Program (MILP). As vantage configurations grow increasingly dense and paths in the network $\mathcal{R}$ approach optimal paths, the MILP solution will approach the optimal solution to the original continuous problem~\eqref{eq:continuous minimization problem}.  However, finding optimal MILP solutions is NP-hard - and finding suitable solutions usually gets harder with the amount of integer variables in the formulation of the problem - which in the case of the formulation in Appendix I scales with $\mathcal{O}(K^2)$. We therefore opt to tackle this problem with a two-stage LP+TSP approach.

The LP first finds an dosage plan, in the form of {\em dwell times} to be spent at each vantage configuration, that is optimal assuming that the robot can instantly ``teleport'' between configurations. Second, the TSP finds the minimum-time traversal of the configurations with non-zero dwell times. Assuming that the robot is sufficiently fast that irradiation is the limiting step, this strategy will produce near-optimal results.

Another issue to be addressed is that the integral in \eqref{eq:Discrete Irradiance} does not have a closed form.  We quickly compute an approximate irradiance vector from every vantage configuration and assemble them into an irradiance matrix using a GPU-based visibility check. 


\subsection{Vantage Configuration Selection}
We first uniformly select a set of light positions in the task space, giving a superset $\{x_1,\cdots,x_{K'}\}$ of $K'$ light positions. For each light position, we solve the inverse kinematics problem for each robot $IK(x_k) = q_k$ and insert $q_k$ into the vantage configuration set if a collision-free IK solution can be found. During IK feasibility computation, the robot's geometry is dilated by 5\,cm to discourage the use of ``coiled`` configurations, since these induce harder planning problems. The selection scheme of $\{x_1,\cdots,x_{K'}\}$ is robot-dependent. If the robot is able to move in 3-D, then they are drawn from an uniform grid in the bounding box of $\mathbb{E}$ in $\mathbb{R}^3$, but if the robot is constrained to 2D motion, like a mobile base, they are drawn from a gridding of the floorplan of $\mathbb{E}$ in $\mathbb{R}^2$.


\subsection{Roadmap Computation}
In this step we compute a PRM \cite{kavraki1996probabilistic} to attempt to connect the vantage configurations $\{q_1,\cdots,q_K\}$ with feasible paths.

The PRM is an undirected graph $R=(V,E)$ consisting of configurations $q\in \mathbb{C}_{free}$, called ``milestones``, and edges $(a,b) \in E$ between milestones $a$ and $b$ are straight line paths that are required to lie completely in the free space, that is, $\overline{ab} \in \mathbb{C}_{free}$. The feasibility of an edge is verified by linearly interpolating between configurations $a$ and $b$ and checking for collisions at regular intervals. In addition, we define the distance between two milestones $a$ and $b$ to be the length of the end-effector trajectory resulting from the linear interpolation between $a$ and $b$ in the workspace.

We then construct $R$ with the following sampling scheme: 

\begin{enumerate}
    \item Add $\{q_1,\cdots,q_K\}$ as initial milestones of the PRM and try to connect pairs of nearby milestones if the edge between them is feasible.
    \item Sample 4 thousand configurations at random from configuration space and attempt to connect them to the existing roadmap $R$. This random sampling is as follows: 30\% of the time, we sample uniformly at random from the configuration space and the remaining 70\% of the time, we sample within a neighborhood of a target milestone, selected uniformly at random.  If at the end of this step all target configurations lie in the same connected component, go to 6; else, go to 3
    \item If the fraction of target milestones within the same connected component ($\phi$) is smaller than 0.8, continue sampling milestones following the procedure in 2 in increments of 200 samples. Otherwise, move to 4
    \item If all target configurations are in the same connected component, go to 6; Else, if $\phi > 0.8$, proceed with targeted sampling. Select one of the target milestones that is yet to be connected to the others uniformly at random, hereby named $q_{focus}$. Find the connected component to which $q_{focus}$ belongs, $\mathcal{C}_{focus}$. Find the connected component containing the majority of the target milestones $\mathcal{C}_{major}$. 
    \item Find the nearest neighbors between $\mathcal{C}_{focus}$ and $\mathcal{C}_{major}$, $q_{focus_{near}}$ and $q_{major_{near}}$. Then, draw samples near either $q_{focus_{near}}$ and $q_{major_{near}}$, uniformly at random. After 10 samples have been drawn in this manner, return to 4. 
    \item Extract the configuration space trajectory between the target configurations from $R$, with no additional processing of the paths (i.e. shortcutting or a-posteriori trajectory optimization), and calculate the distances between them.
    
\end{enumerate}

Step 6 is introduced to help the planner focus its sampling on narrow passages in configuration space. 

After $R$ is computed, vantage points that are not in the largest connected component are discarded.  For the remaining points, the shortest paths in $R$ between all pairs $(q_k,q_l)$ are computed to form the distance matrix $d_{kl}$. Note that one consequence of this targeted sampling approach is that milestones that lie in free space tend to be connected early on and, thus, tend to have more jagged paths between them. 

\subsection{Discrete Radiative Fluence}
We approximately calculate the radiative fluence matrix with entries $I_i(x_k)$. Note that a typical environment in 3D contains millions of triangles ($N$) and we will sample tens of thousands of potential vantage configurations ($K$). Therefore, the matrix size $I_i(x_k)$ is huge and its calculation can typically become the bottleneck. We provide a GPU-based implementation that can calculate each column $I_*(x_k)$ in milliseconds.

\begin{figure}[tbp]
\centering
\includegraphics[width=0.40\textwidth]{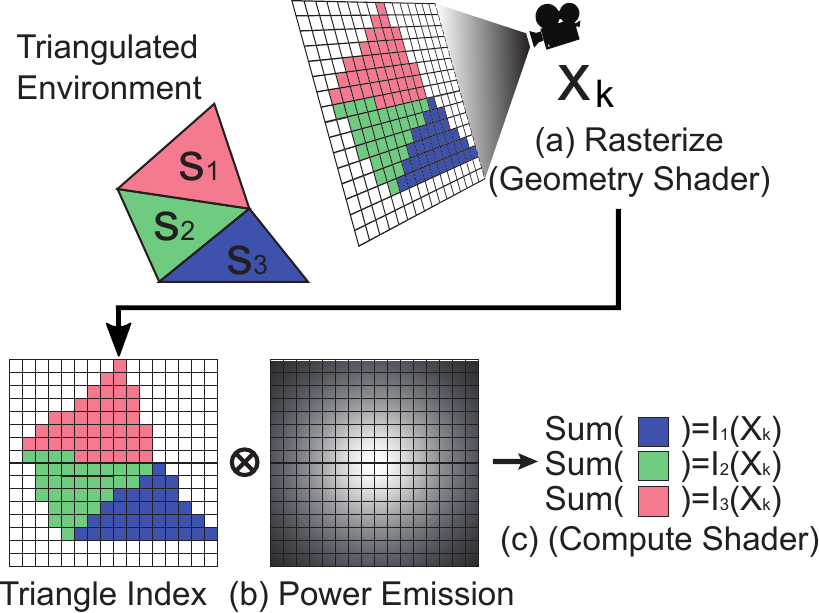}
\caption{\label{fig:Irradiance} Illustrating the GPU-based irradiance calculation. (a): The triangle index is rasterized into an environment map using geometry shader. (b): The power emission $e(i,j)$ is precomputed. (c): The operation $F[T[i,j]]$ += $e(i,j)$ is performed using hardware accelerated pixel-blending. (Best seen in color)}
\vspace{-10px}
\end{figure}

The irradiance is a measure of the rate of radiant exposure, and is given in the units of watts per square meter. We first describe the simple case where the robot is a point light source, i.e. $\E{X}=\mathbb{R}^3$. We assume that reflected light is not a major source of illumination, so that the irradiance density received by the infinitesimal patch $ds$ is given according to the inverse square law:
\begin{equation}
\label{eq:inverse square law}
\left\langle\mathbb{I}(x_k,ds),n(s)\right\rangle =
\begin{cases}
0    & \text{$ds$ visible from $x_k$} \\
\frac{P\langle s-x_k,n(s)\rangle }{4\pi\|s-x_k\|^3} & \text{otherwise,} 
\end{cases}
\end{equation}
where $P$ is the power (or radiant flux) of the light source and $s$ is the location of the infinitesimal surface patch. A patch is considered visible only if $\left\langle y-x_k,n(s)\right\rangle>0$ and no other surface lies closer to $x_k$ along the ray $y-x_k$.

If no other triangles are in the way from $x_k$ to the entire triangle $s_i$, then the irradiance can be calculated according to \cite{mosher1999eeg}, i.e. the integral of \prettyref{eq:Discrete Irradiance} has closed form solution. However, when occlusion occurs, no closed form solution can be found for the per-triangle irradiance. Instead, our GPU-based implementation calculates the irradiance $I_i(x_k)$ by rasterization. This roughly follows the pipeline for radiosity calculations used in computer graphics~\cite{cohen1985hemi,coombe2004radiosity} disregarding Lambertian reflectance. Our implementation (Fig.~\ref{fig:Irradiance}) is comprised of the following steps:
\begin{itemize}[leftmargin=*]
\item The scene is rasterized using a standard graphics pipeline, with the camera centered at $x_k$. Each triangle's index is rendered into the pixel buffer $T$ bound to a cubemap texture (the {\em visibility cube}) using framebuffer object and a geometry shader \cite{green2005opengl}. In the meantime, a Z-buffer is used for visible surface determination. After rasterization, we store the value $T[i,j]$ for each pixel $(i,j)$ on the image plane. $T[i,j]$ is the index of the closest triangle intersecting the ray from pixel $(i,j)$ to $x_k$. A void pixel indicates that no triangle is occupying the pixel.
\item For each pixel $T[i,j]$ containing a visible triangle, the amount of power $e(i,j)$ emitted over the solid angle subtended by the pixel is calculated using \cite{mosher1999eeg} and all power terms $e(i,j)$ belonging to $T[i,j]$ are summed up and stored in the triangle buffer $F$. This summation of $e(i,j)$ is performed using the GPU's hardware accelerated pixel-blending. In particular, we first set the triangle buffer $F$ as the render target and store $T[i,j]$ in the GPU buffer. We then execute a shader program for each $e(i,j)$, where we check $T[i,j]$ for the index in $F$ and use geometry shader \cite{bailey2016opengl} to render a single pixel into $F$, with color equal to $e(i,j)$ and pixel-blending turned on. The accumulated value for each triangle is the radiant flux, which measures irradiance integrated over the non-occluded area of the triangle. 
\item The radiant flux $F[i]$ is divided by the area of each triangle to obtain the mean irradiance $I_i(x_k) = F[i]/|s_i|$.
\end{itemize}
Because this process will be performed repeatedly, the power emission $e(i,j)$ for each pixel is precomputed and stored in a separate texture of the same dimensions as the rendered buffers, denoted as $E$, so that it can be retrieved with a single memory lookup. A note-worthy caveat of our method is the use of mean irradiance $I_i(x_k) = F[i]/|s_i|$ to replace the true uneven irradiance distribution within a single triangle, which can be remedied by having more finely discretized meshes. 

{\em Non-Point Light Sources:} Our procedure to compute $I_i(x_k)$ can be naturally extended to non-trivial light source shapes, such as an omnidirectional cylindrical light source. In those instances, the surface of light sources can be approximated by a set of evenly distributed point sources, where each point source emits an equal fraction of the light's total radiant power. The total radiant flux is accumulated for each point before dividing by the area of each triangle to obtain the irradiance. More advanced shader programs such as \cite{heitz2016real} can also be used to approximate the continuous integration of light contributions along the light source's surface area on GPU. For light sources with uneven irradiance distribution, such as shielded or mirrored lights, we can replace the power emission texture $E$ with a precomputed custom distribution. 

If the light source is not standalone but mounted on a robot, then the position of the light source $p$ is determined by its forward kinematics, which is denoted as $p(x_k)$ and plugged into \prettyref{eq:inverse square law} in the place of $x_k$, arriving at $\mathbb{I}(p(x_k),n(s))$.

\subsection{\label{sec: Sequential LP}Approximate Two-Stage Optimization}
At this point, all related variables of \prettyref{eq:discrete general trajectory minimization} have been calculated. The first stage proceeds by relaxing all $z_{kl}=1$ and derives an optimal set of dwell times. Assuming no transit time, the optimal dwell times can be determined by solving the following linear program:
\begin{equation}
\begin{aligned}
\label{eq:LP}
\argmin{t_k}&\sum_{k=1}^Kt_k \\
\ST&\mu_i \geq \mu_{min}\quad\forall i=1,\cdots,N,
\end{aligned}
\end{equation}
A potential issue with \prettyref{eq:LP} is that it does not account for partially infeasible problems, which frequently occur in practice because some triangles $s_k$ are totally invisible from all vantage configurations. In these cases, \prettyref{eq:LP} will report infeasibility rather than return an approximate solution.  To remedy this problem, we propose the following relaxed LP that always returns feasible solutions:
\begin{equation}
\begin{aligned}
\label{eq:simple LP}
\argmin{t_k,\sigma_k\geq0}&\sum_{k=1}^Kt_k+\sum_{i=1}^Np_i\sigma_i \\
\ST&\mu_i+\sigma_i \geq \mu_{min}\quad\forall i=1,\cdots,N  \\
&\sum_{k=1}^K t_k \leq T_{max},
\end{aligned}
\end{equation}
where $p_i$ denotes the infeasibility penalty of a triangle $s_i$ and $\sigma_i$ is a slack variable allowing all constraints to be satisfied in the worst case. We further constrain the time budget for disinfection to $T_{max}$. With large penalties $p_i>\|I_*(x_*)\|_F$ and sufficiently large $T_{max}$, the solution to the LP tends to set all $\sigma_i=0$ and the solution to \prettyref{eq:simple LP} approaches the solution to \prettyref{eq:LP}. When some surfaces are totally invisible or disinfection cannot be accomplished within the time budget, the LP solution accepts $\sigma_i=\mu_{min}-\mu_i>0$ for some indices $i$, thereby accepting penalty $p_i (\mu_{min}-\mu_i)$. For prioritized surface patches $s_i$, a larger $p_i$ should be used so the LP tends to avoid positive $\sigma_i$.  To solve \prettyref{eq:simple LP} we leverage the large-scale interior-point algorithm implemented in Gurobi~\cite{gurobi}.

The second stage in this approximate approach solves the TSP problem to find a tour amongst vantage points with nonzero dwell times, that is, minimizes transit times amongst edges $\{z_{kl}|t_k>0\wedge t_l>0\}$. While this problem is still NP-hard, it is solved over a much smaller set of candidate paths. In addition, since it fits the traditional TSP formulation, we are able to leverage polynomial-time approximate TSP solvers, such as \cite{Helsgaun2000EffectiveHeuristic}, which have near-optimal performance for relatively small euclidean TSP instances as the ones we encounter. Once the tour is found, the final disinfection trajectory is obtained by linearly interpolating in configuration space along the edges of the roadmap. 
\section{Experiments}

Our experiments aim to answer the following questions:
\begin{enumerate}
    \item How much better is the coverage of an optimally planned disinfection trajectory if compared to a single-point strategy?
    \item How large is the optimality penalty incurred by solving the problem sequentially vs using an optimal MILP formulation?
    \item How do different robot designs compare in terms of maximum disinfection coverage and efficiency?
\end{enumerate}
We use a simplified 2.5D experiment to test questions 1 and 2, and a realistic environment in 3D for question 3.  

In each experiment, all surfaces require a minimum disinfection fluence $\mu_{min}=280$ \,J/m$^2$, which is a conservative estimate of the necessary fluence to induce a $3\log_{10}$ reduction in infectivity of SARS-Cov2 \cite{Heling2020UltravioletStudies.}. In addition, the light is assumed to have a constant radiant flux, P, of 80\,W and that the maximum speed of all robot end-effectors is 0.5\ m/s.

\subsection{Comparison with static illumination}
\begin{figure}[tbp] 
\vspace{-3px}
\begin{subfigure}{0.24\textwidth}
\includegraphics[width=\linewidth]{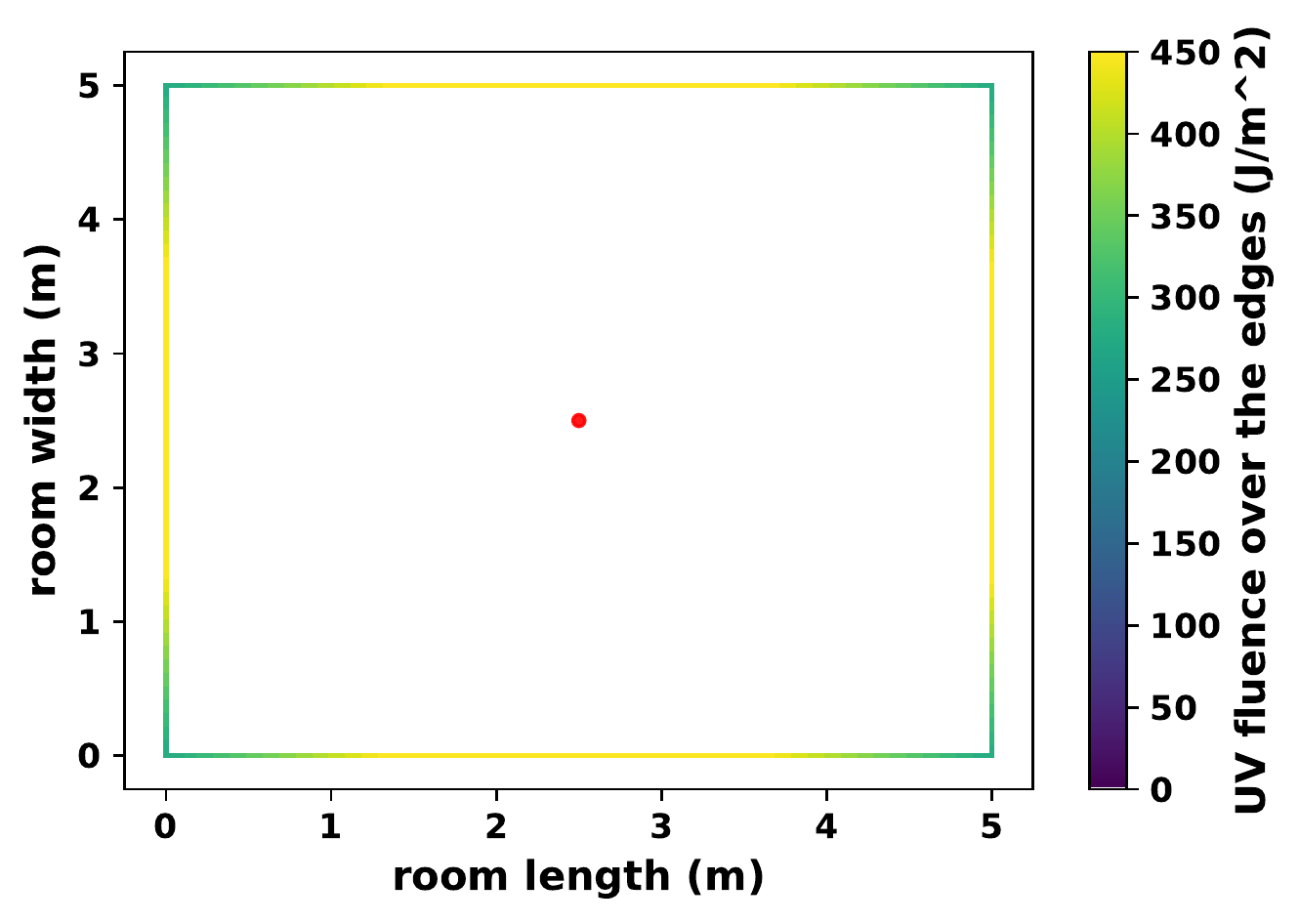}
\caption{Stationary illumination}
\end{subfigure}
\begin{subfigure}{0.24\textwidth}
\includegraphics[width=\linewidth]{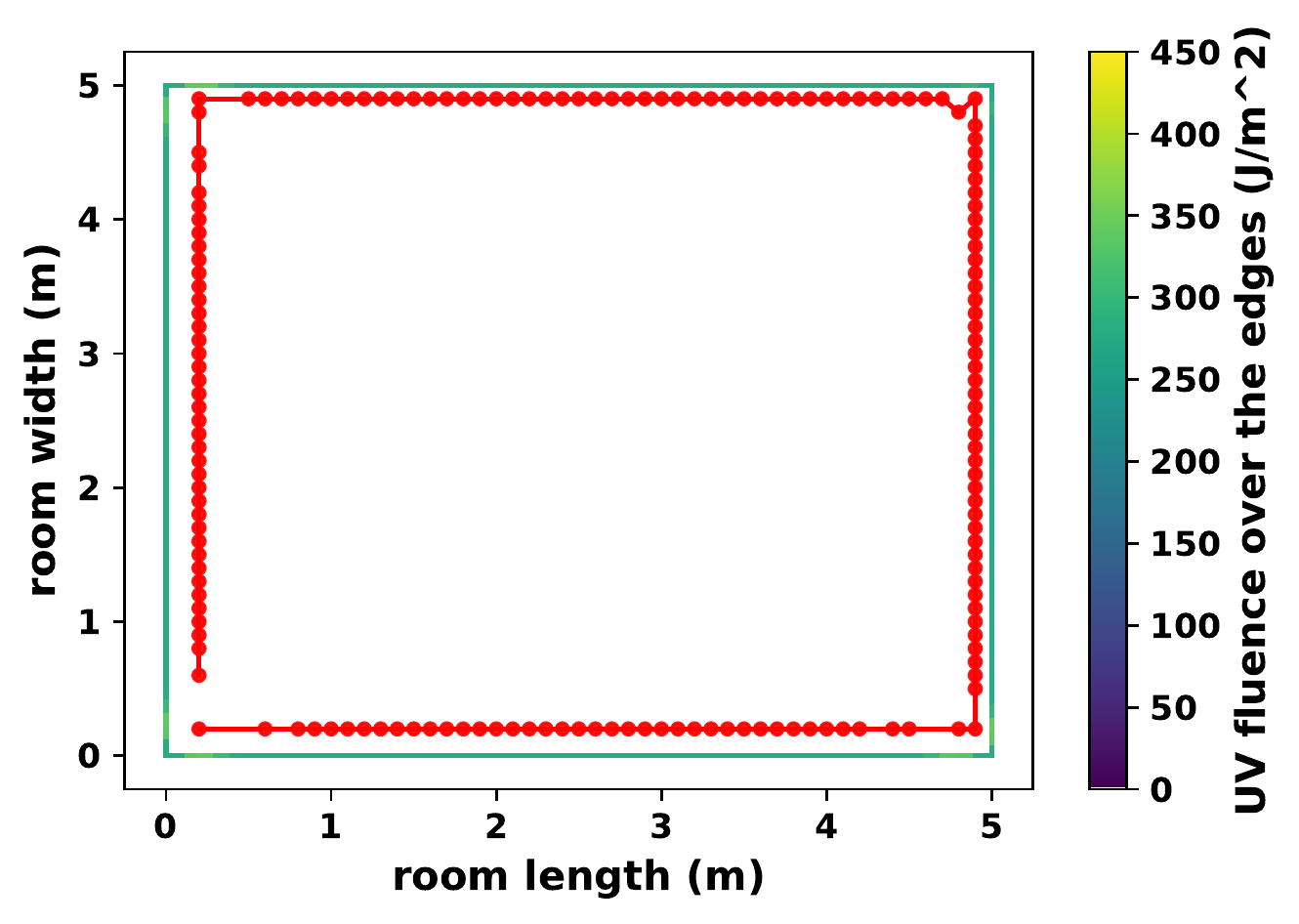}
\caption{Our method} 
\end{subfigure}
\vspace{-3px}
\caption{Empty room disinfected by the best stationary point (red dot) and by our method. Each surface is colored by its received fluence, and the optimized trajectory is drawn in  red. (Best seen in color)}
\label{fig: single point vs multi-point 2.5D empty}
\vspace{-10px}
\end{figure}

First we evaluate disinfecting the walls of a 5m$\times$5m empty room as a 2.5D problem. Walls are 2\,m meters tall and a spherical point light source is used. We consider a discretized version of the room where each wall is subdivided into fixed-length subsegments, and irradiance from a point can be calculated analytically for rectangles~\cite{4121581}.  We consider a static illumination strategy that places the disinfection light to have maximum coverage over the obstacle space, allowing it to irradiate the surfaces for as long as necessary to fully disinfect its visible surfaces. In our method, we treat the robot as a cylindrical base of radius 10 cm, and constrain the movement of the light to a plane at height 1\,m.  Vantage points are sampled along a 0.1\,m grid. The static method takes 143.7 minutes to reach full room disinfection, while ours does so in 95.6 minutes, including movement time between vantage points - the contrast between solutions is illustrated in \prettyref{fig: single point vs multi-point 2.5D empty}

\begin{figure}[tbp] 
\vspace{-3px}
\begin{subfigure}{0.24\textwidth}
\includegraphics[width=\linewidth]{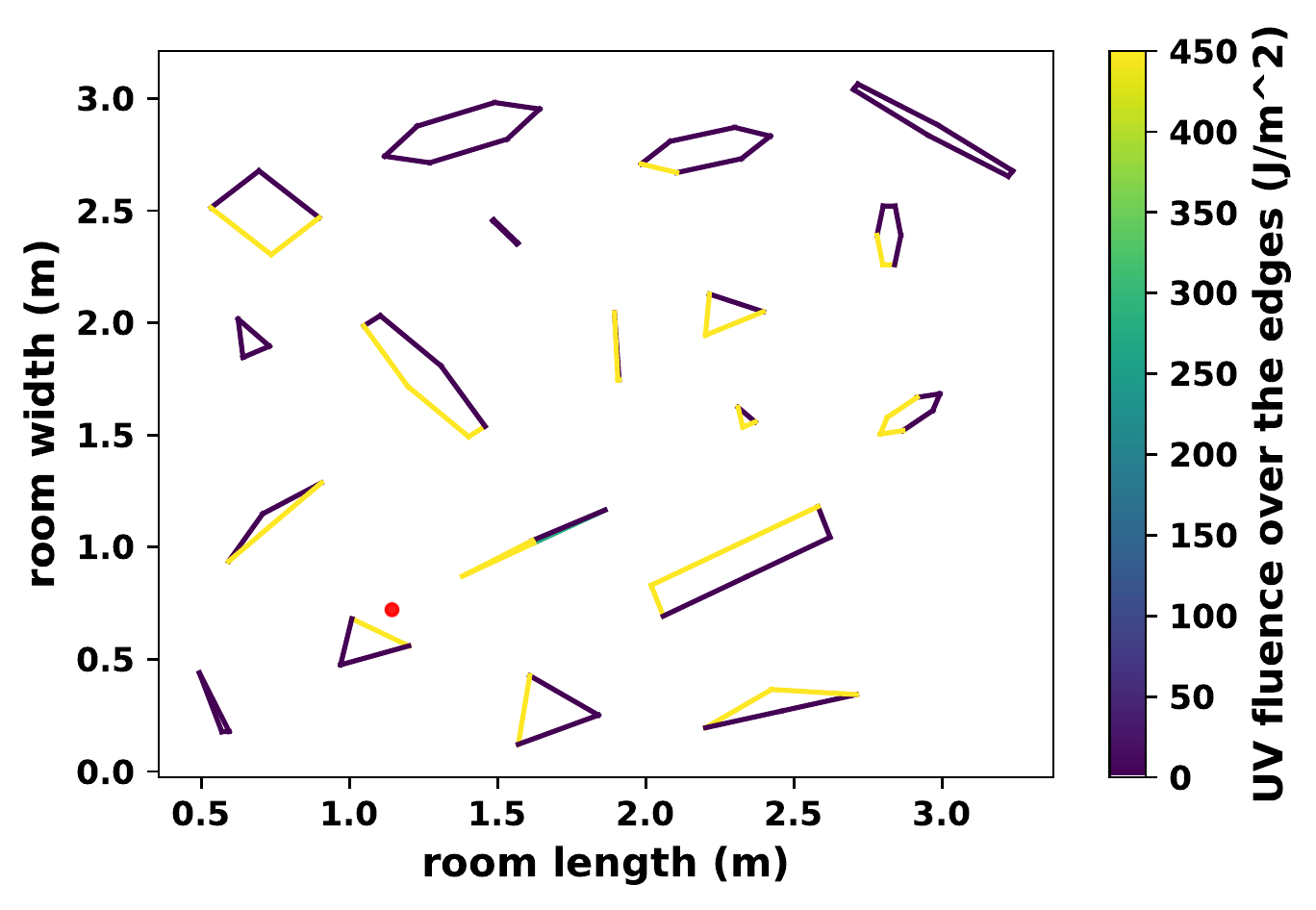}
\caption{Stationary illumination}
\end{subfigure}
\begin{subfigure}{0.22\textwidth}
\includegraphics[width=\linewidth]{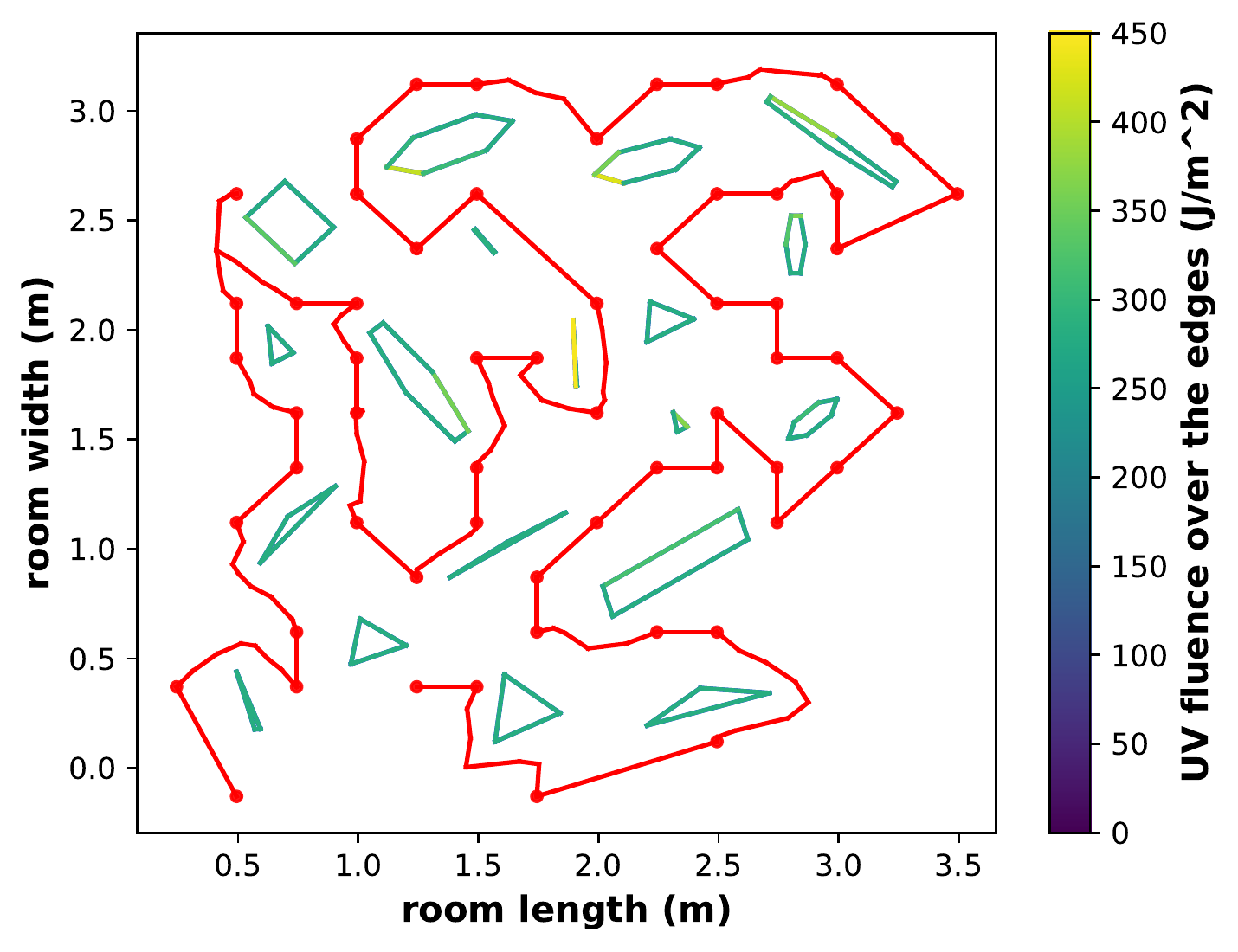}
\caption{Our method}
\end{subfigure}
\vspace{-3px}
\caption{Same room disinfected by the best stationary point (red dot) and by our method. Each surface is colored by its received fluence and the trajectory is drawn in red. (Best seen in color)}
\label{fig: single point vs multi-point 2.5D}
\end{figure}

\begin{figure}[tbp] 
\centering
\includegraphics[width=0.8\linewidth]{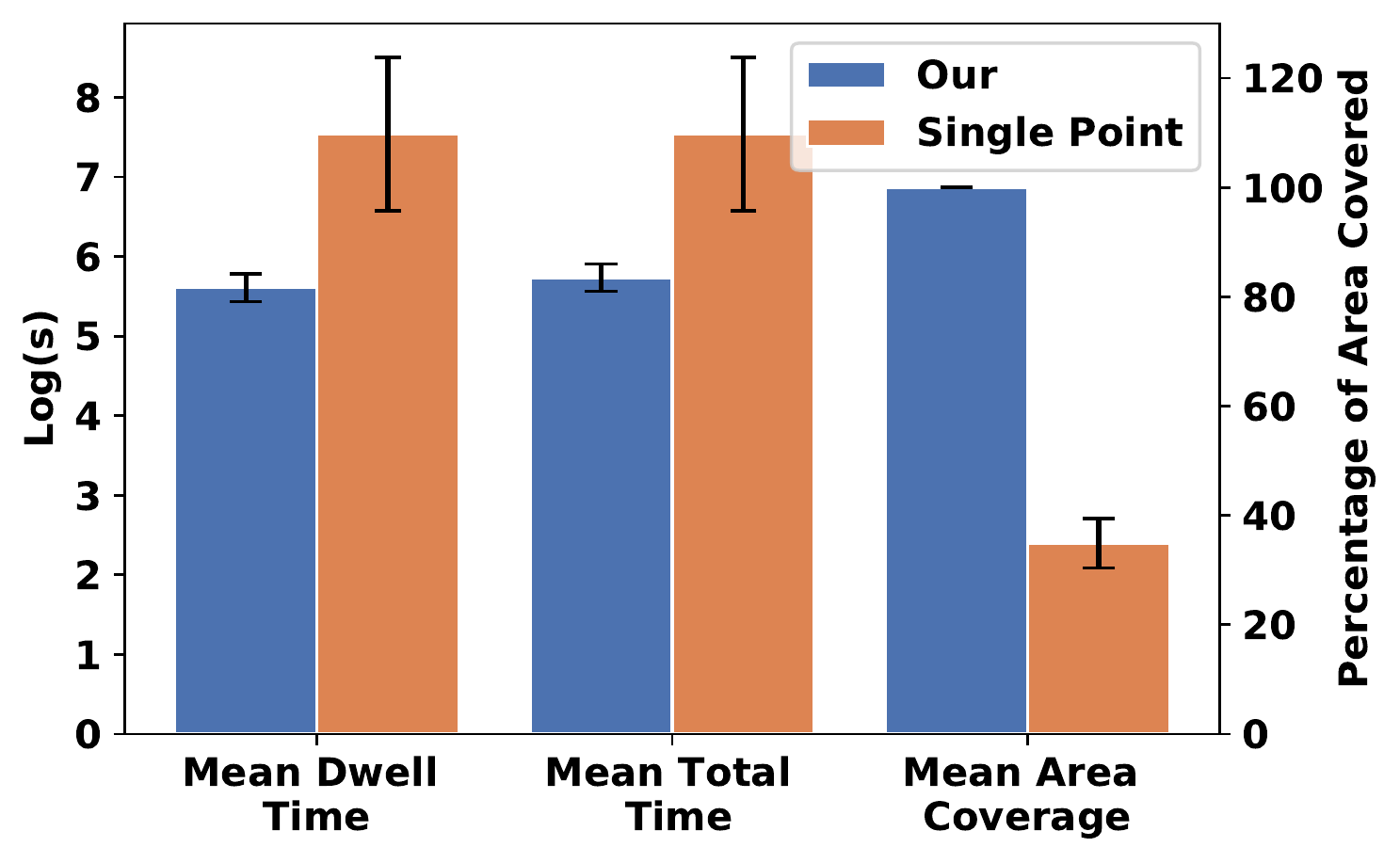}
\caption{2.5D Disinfection performance comparisons with best single point strategies} \label{fig: Single point 2.5 D comparisons}
\vspace{-12px}
\end{figure}

Next, we randomly generate 25 2.5D rooms in a 4\,m$\times$4\,m area and with 2\,m tall polygonal obstacles. Each world contains a random number of obstacles (between 7 and 19), with each obstacle randomly generated by scaling, shearing and displacing regular polygons. Visibilities of each segment from a given vantage point are determined by creating a visibility graph amongst vantage points and segment midpoints \cite{lozano1979algorithm}.   \prettyref{fig: single point vs multi-point 2.5D} shows the output for one example. Note that for our method, all segments are covered, and few segments are overexposed.
\prettyref{fig: Single point 2.5 D comparisons} shows results averaged over all rooms, indicating that our proposed method consistently disinfects 100\% of the environment, whereas the optimal static illumination only disinfects 35\%.  Moreover, to disinfect the visible segments, static illumination requires approximately 2 orders of magnitude more time.

\subsection{Comparison against MILP formulation}
\begin{figure}[tbp] 
\centering
\includegraphics[width=0.8\linewidth]{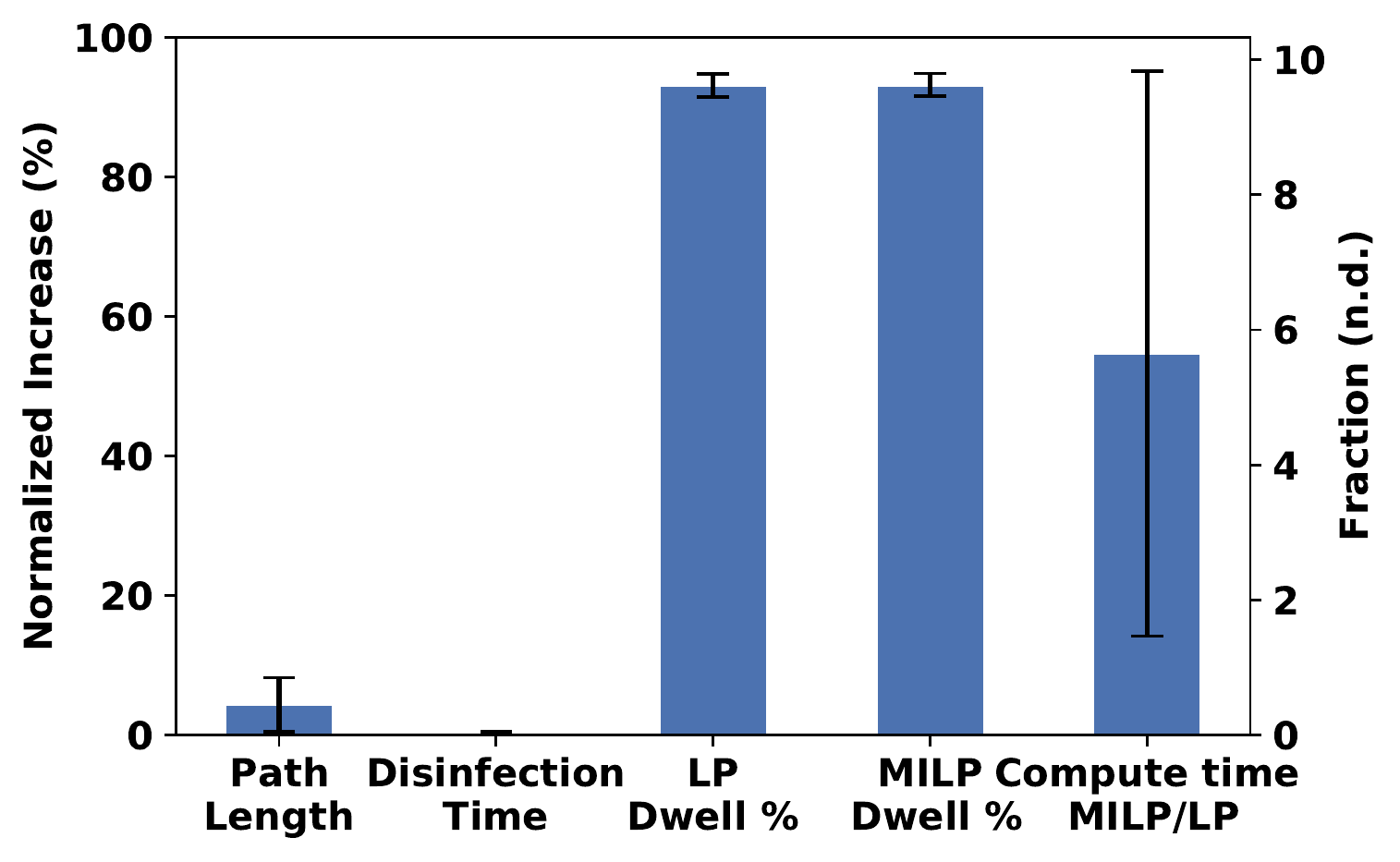}
\caption{2.5D performance comparisons between MILP and LP+TSP Formulations} \label{fig: LP and MILP Comparisons}
\vspace{-12px}
\end{figure}

Next, we compare the two-stage scheme and the globally optimal MILP formulation described in our extended report~\cite{extended_report}. We limited the point robot to travel along a grid with 0.5m spacing, resulting in $\approx$ 64 candidate vantage points and $\approx$ 4k boolean variables in the MILP(to ensure MILP feasibility within constrained time). These results are illustrated in \prettyref{fig: LP and MILP Comparisons}.  Columns 1, 2 indicate the mean percentage difference between MILP and LP solutions; columns 3 and 4 indicate the mean percentage of dwell time w.r.t. the total disinfection time in each solution and column 5 is the mean value of the ratio between the computation time of the MILP over the computation time for the LP+TSP. Note that path lengths differ, on average, less than 10\% between the two solution methods - which is evidenced by the mean difference of less than 5\% in total disinfection time. In addition, we verify our initial assumption that dwell times make up the majority of disinfection time, since they represent over 90\% of the total disinfection times in both cases. Moreover, the two-stage approach is $6$ times faster, on average, to compute than the optimal MILP solution - even in examples with coarse grid. Note that in a grid with twice the resolution ($\approx$  256 candidate vantage points - $\approx$  65k boolean variables in the MILP), we could not get solutions  within 2 hours for the MILP formulation, while the LP+TSP solutions took about 5 seconds to compute, on average. Additional experiments evaluating the impact of grid resolution and environment discretization resolution can be found in Appendix II.




\subsection{Comparing Robot Designs}
\begin{figure*}[tbp] 
\centering
\setlength{\tabcolsep}{20px}
\begin{tabular}{cc}
\includegraphics[width=.4\linewidth]{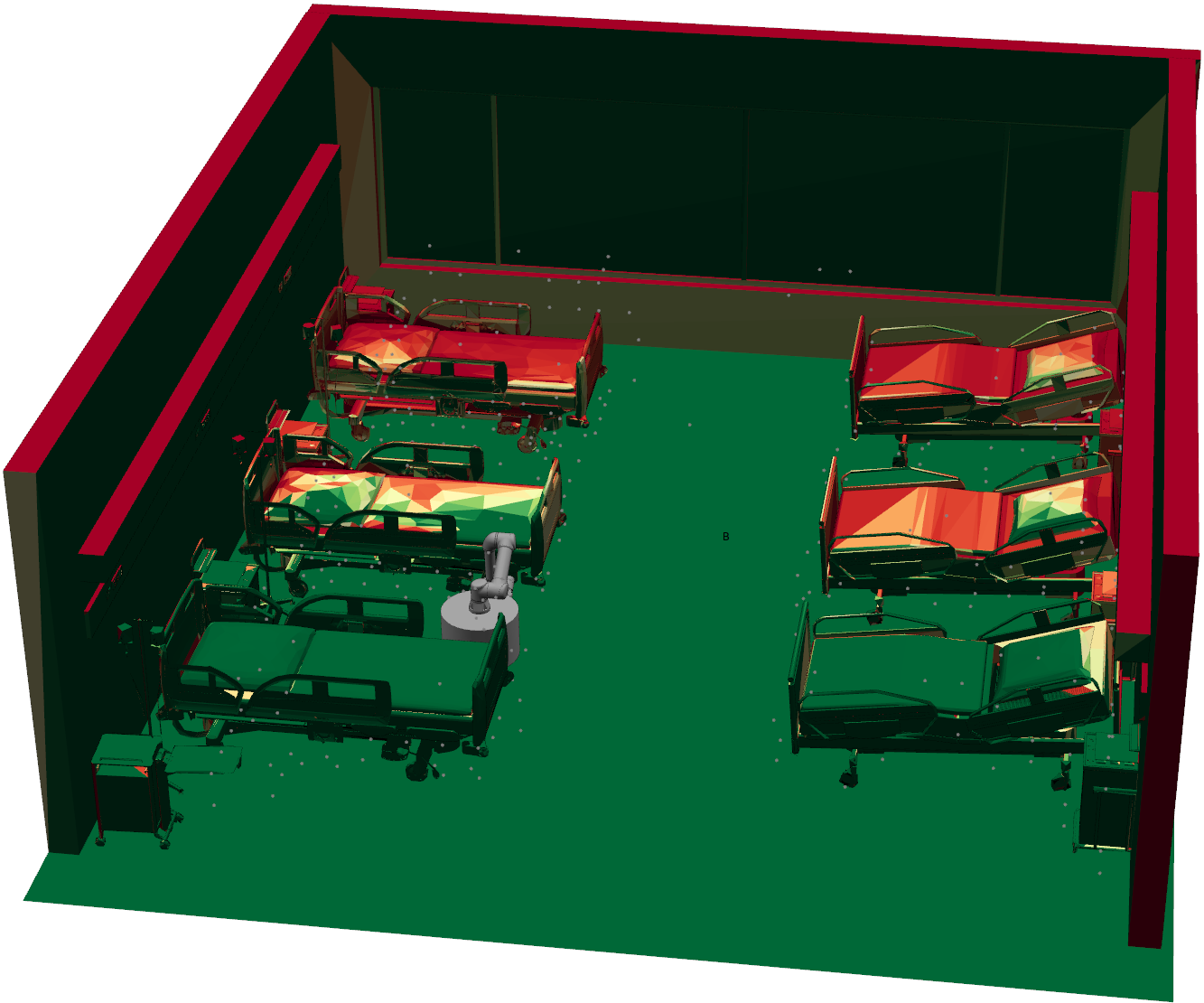}
\put(-110,0){\small{7.5\,min}} &
\includegraphics[width=.4\linewidth]{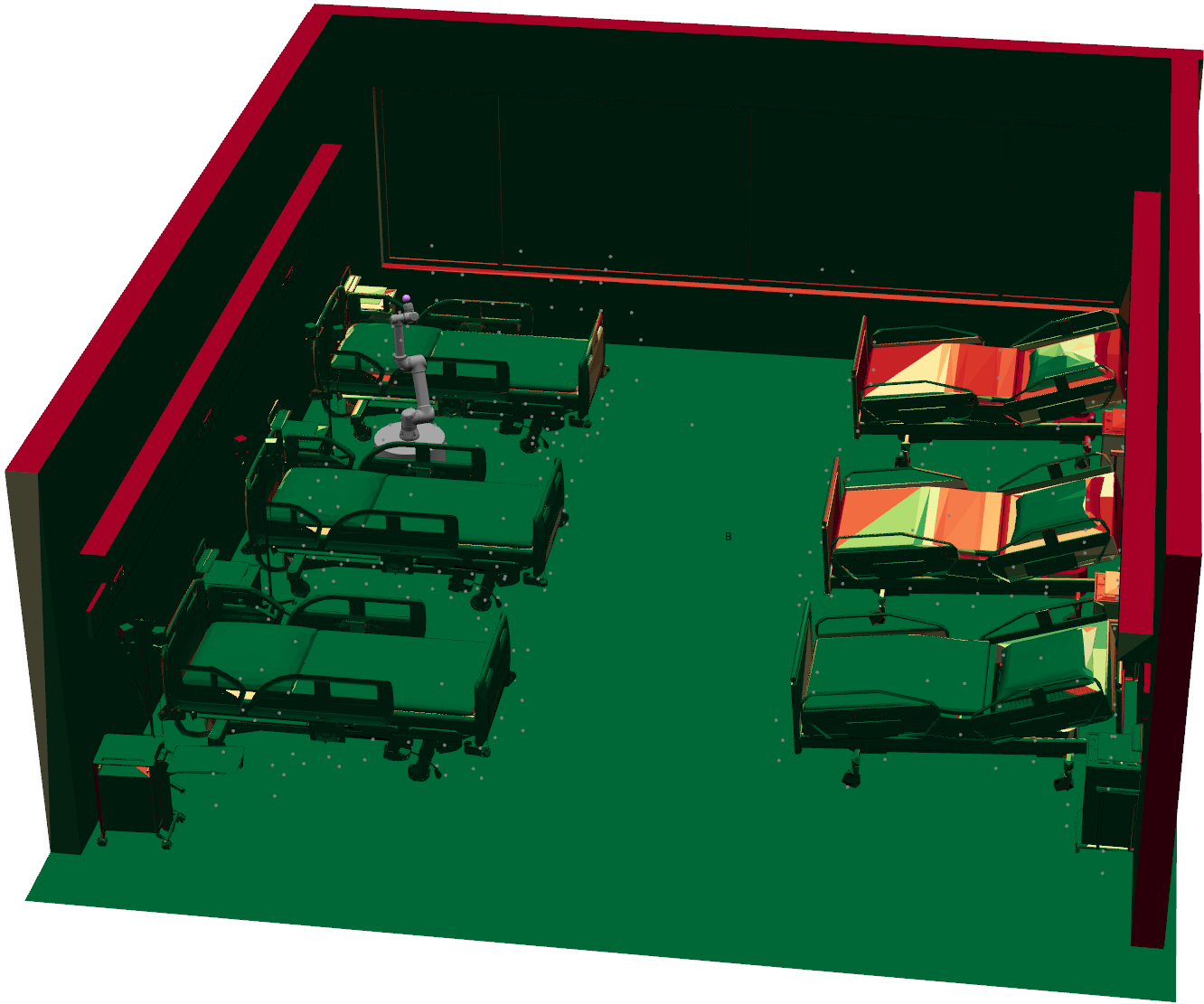}
\put(-110,0){\small{15\,min}}\\
\includegraphics[width=.4\linewidth]{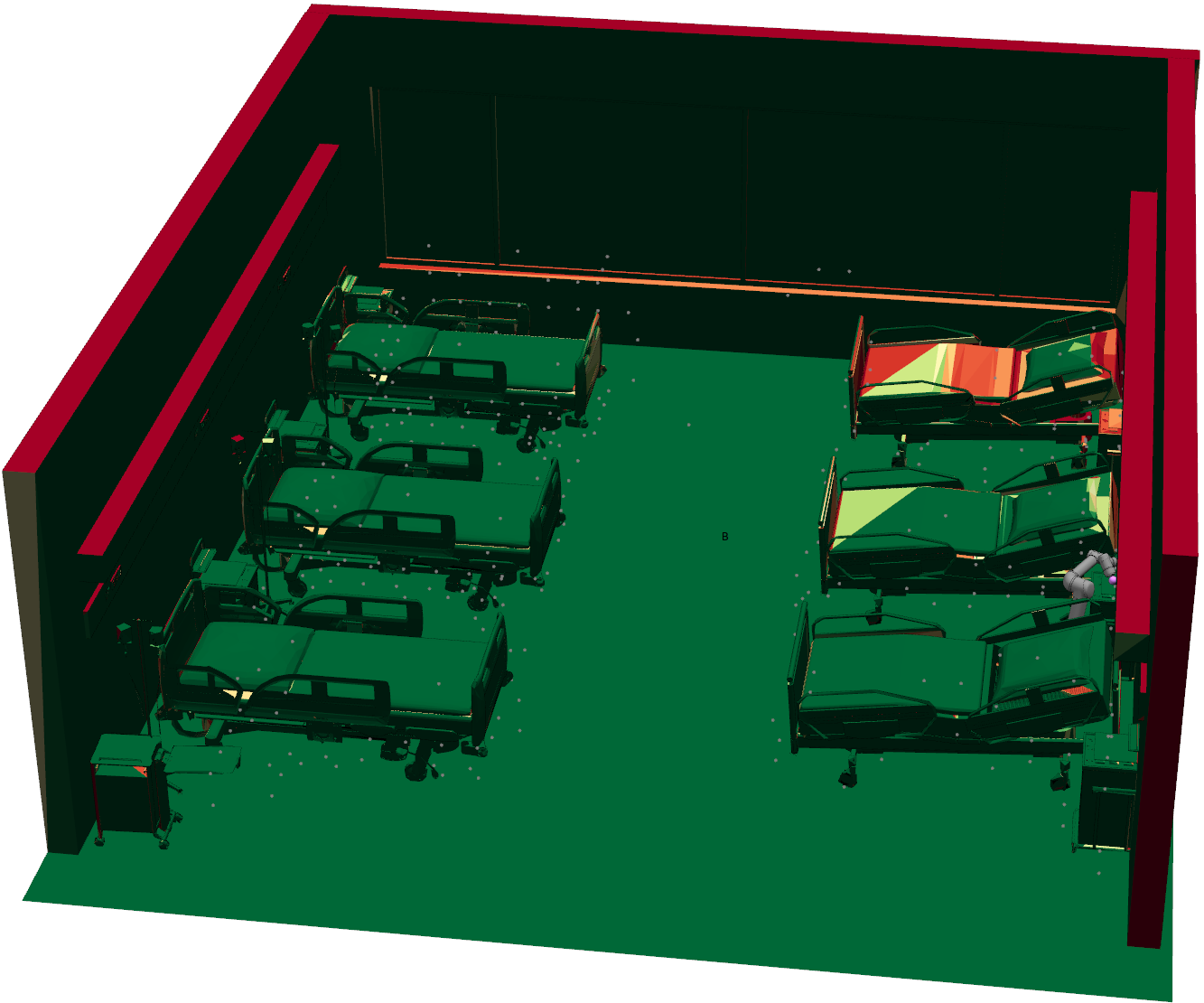}
\put(-110,0){\small{22.5\,min}} &
\includegraphics[width=.4\linewidth]{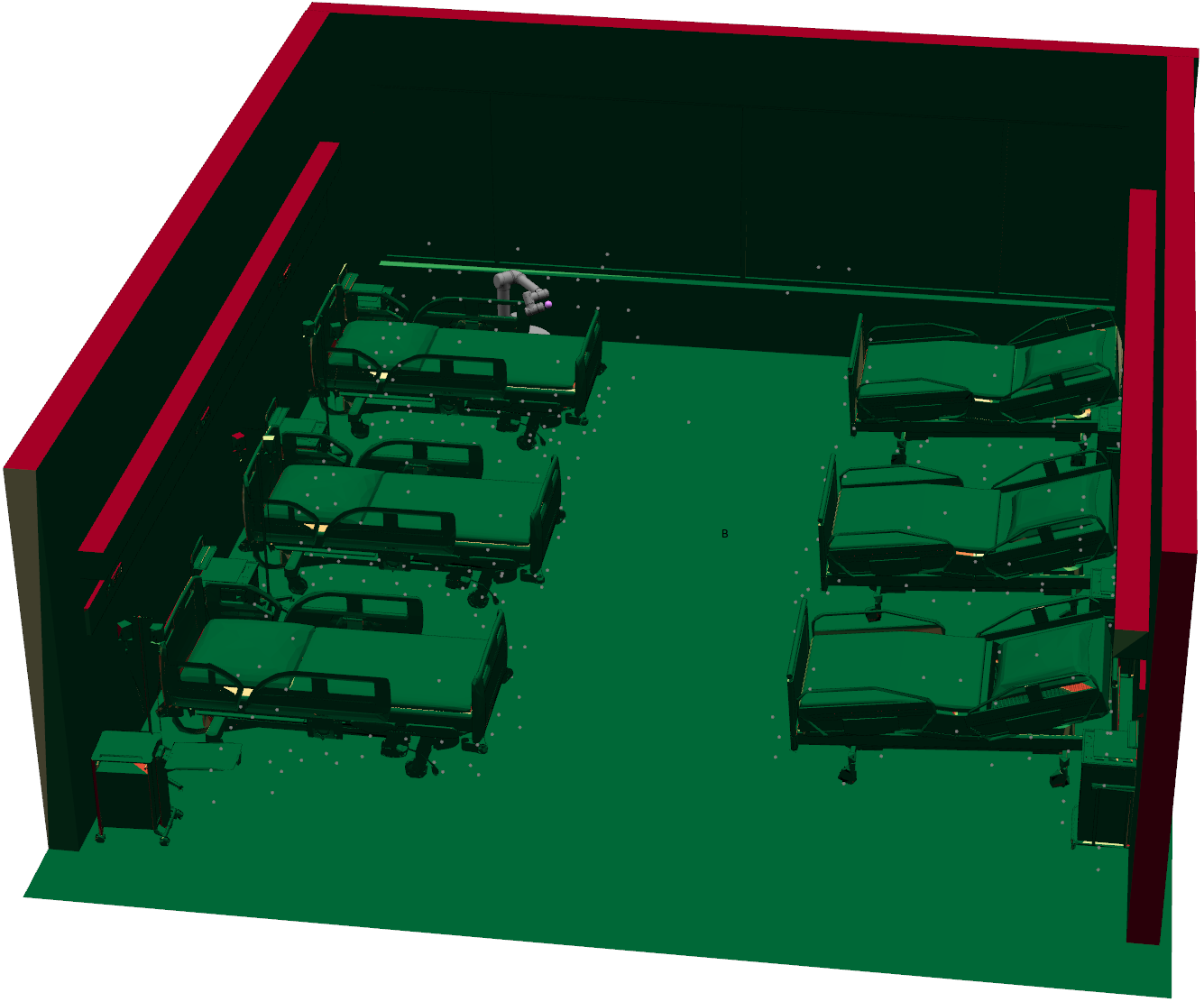}
\put(-110,0){\small{30\,min}} 
\end{tabular}
\caption{Time lapse of Armbot's disinfection progress for an infirmary within a time budget of 30 minutes. (Best seen in color)} \label{fig:Smart 3D disnfection}
\end{figure*}

Our 3D tests were performed in a hospital infirmary's CAD model\footnote{https://grabcad.com/library/hospital-ward-2-2} (\prettyref{fig:Smart 3D disnfection}), simplified to 60k triangles using quadric edge collapse decimation. 
The method described in Sec. IV-C is configured to use 512x512 resolution framebuffers for computing irradiances.

\begin{figure}[tbp] 
\centering
\includegraphics[width=0.8\linewidth]{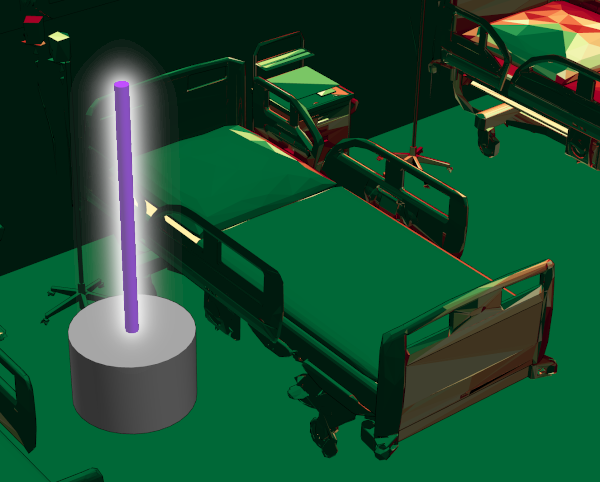}
\caption{Towerbot mid-disinfection} \label{fig: towerbot mid-disinfection}
\vspace{-12px}
\end{figure}

\begin{figure}[tbp] 
\centering
\includegraphics[width=0.8\linewidth]{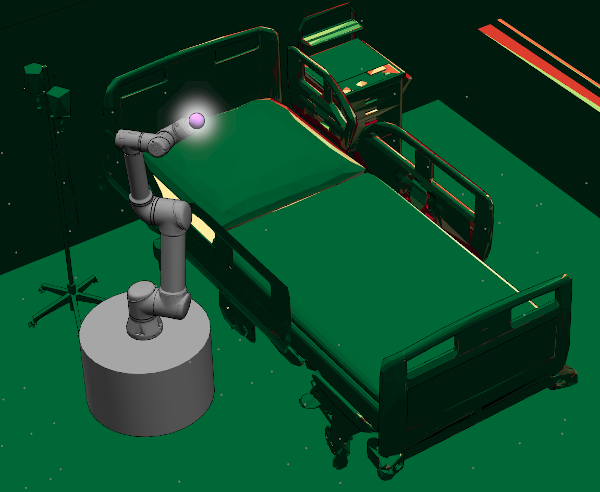}
\caption{Armbot mid-disinfection} \label{fig: armbot mid-disinfection}
\vspace{-12px}
\end{figure}

We compare three models for the disinfection robot: ``Floatbot``, a freely-moving spherical light source, ``Towerbot``, a cylindrical mobile base that moves in the plane, is 55\,cm in diameter and 37\,cm tall, and has a 1.2\,m tall cylindrical light source attached to its top, seen in (\prettyref{fig: towerbot mid-disinfection}); and ``Armbot``, a mobile base upon which a UR5e 6-DOF manipulator is mounted and holds a spherical point light source, seen in (\prettyref{fig: armbot mid-disinfection}), with their lamps highlitghted.
Floatbot is an idealized model of maximum performance. Towerbot is a model for commercially available mobile disinfection robots (like the Akara Violet and UVD robot's Model B and C) \footnote{https://www.uvd-robots.com/robots | https://www.akara.ai/violet.html}, while Armbot represents a potential advancement that can access more hard-to-reach surfaces than a tower design. All solutions were computed within 50 minutes, with the irradiance matrix calculation taking over 80\% of the time on all 3D experiments.


\begin{figure}[tbp] 
\centering
\includegraphics[width=0.8\linewidth]{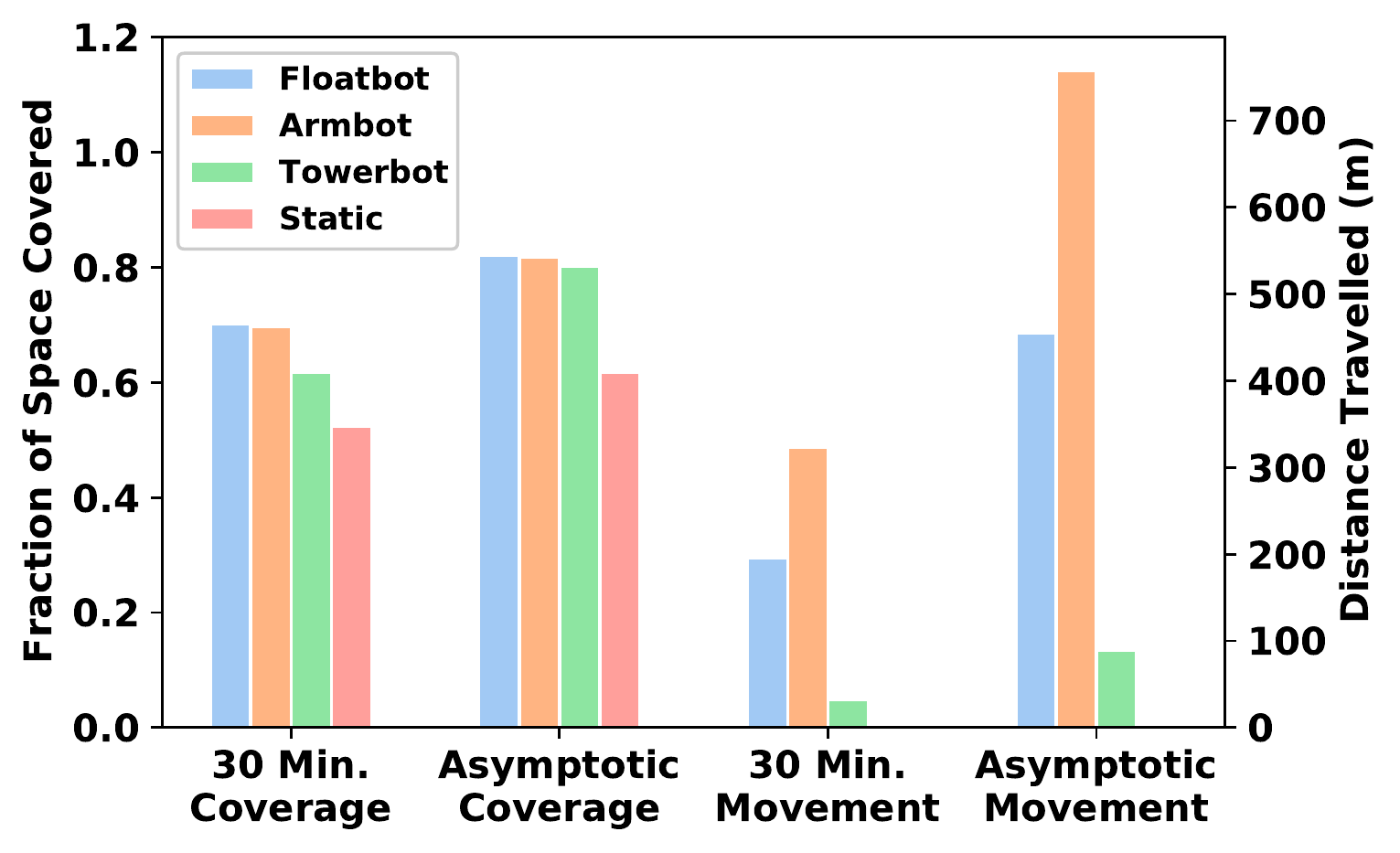}
\caption{Performance of robot designs disinfecting an infirmary. } \label{fig: Robot Design Comparisons}
\vspace{-12px}
\end{figure}

Our experiments designate an irradiation time limit of $T_{max}=30$ minutes and 100 hours for evaluating asymptotic performance. For vantage point selection, we define a 3D grid with resolution 0.25\,m (resulting in 8547 vantage candidates).  We also compare with the strategy of placing Towerbot in the center of the room for the prescribed time budgets to mimic the static status-quo. During motion planning, 4k feasible samples are drawn to create the PRM (with additional samples added in increments of 20 if full milestone connectivity is not achieved).

Results are shown in \prettyref{fig: Robot Design Comparisons}.  We find that robots with more freedom to explore the free space, like Armbot and Floatbot, disinfect a larger area under a given time budget.  Matching experimental results,  \cite{Lindblad2019Ultraviolet-CNeeded}, the status-quo of stationary placement of Towerbot fails to cover much of the surface, as illustrated in \prettyref{fig: single point 3D disnfection}. The asymptotic performance is nearly identical among all mobile solutions, whereas the disinfection efficiency comes with a tradeoff in total distance travelled, among which Towerbot has the smallest trajectory length and Armbot has the longest. This is presumably due to two factors. First, distances in higher dimensions tend to be higher (3D vs 2D) and, second, motion planning for Armbot involves many steps that are prone to sub-optimality, such as vantage configuration selection given a desired lamp position and high-dimensional multi-query path planning. Floatbot's trajectory length is a trivial lower bound to Armbot's trajecory length. More details about the trajectories can be found the attached suplemental video. Additional experiments evaluating the effect of grid resolution on the disinfection performance of the 3 robots can also be found on Appendix II.

\section{Conclusion \& Future Work}
We presented a targeted approach to solve coverage planning problems for UV light disinfection. Our optimization minimizes the disinfection time while ensuring maximum coverage by imposing constraints of minimal irradiance exposure of surfaces. We show that globally optimal solutions can be found by solving a NP-hard MILP and propose a two-stage approximation scheme that can find near optimal solutions with less than $3\%$ sacrifice of optimality while being orders of magnitude faster. We also confirm real-world experiments \cite{Lindblad2019Ultraviolet-CNeeded} that show limitations of stationary UV disinfection robots. Furthermore, our algorithm is general enough to analyze different robotic disinfection mechanisms.
Code for the method is available at \href{https://github.com/joaomcm/Optimized-UV-Disinfection}{https://github.com/joaomcm/Optimized-UV-Disinfection}.
In future work, we would like to analyze the MILP formulation and its interaction with the continuous path planning component. Second, we hope to test the proposed pipeline in a physical system to evaluate how positioning errors from SLAM algorithms and reconstruction errors affect disinfection performance. Third, our vantage configurations are sampled along a uniform task space grid, which may not be the most efficient choice. Finally, we would like to study how joint optimization of vantage configurations, task-space points, and paths could yield more efficient traversals. 

\vspace{12pt}
\ifarxiv
\clearpage
\appendices
\section{Mixed-Integer Linear Programming}
It is evident from \prettyref{eq:discrete general trajectory minimization} that we have a discrete combinatoric optimization problem - which aims to minimize the time needed to disinfect a surface by jointly optimizing the sum of dwell times and the time taken to move between the vantage points $x_k$. However, there are two consistency constraints, the mathematical form of which is unclear. In this section, we reduce these constraints to a set of mixed-integer linear equations. 

The second consistency constraint requires that $t_k$ can only be non-zero when some $z_{kl}=1$ or $z_{lk}=1$. These constraints can be re-written using a big-constant $M$ as follows:
\begin{align}
\label{eq:bigM}
0\leq t_k\leq M\sum_{l=1}^K(z_{kl}+z_{lk}).
\end{align}
The first consistency constraint requires that $z_{kl}$ form a simple path. This constraint can be reduced to mixed-integer linear equations using a similar technique as mixed-integer re-formulation of the Traveling Salesman Problem (TSP). The only difference is that we are not requiring the salesman to pass through every vertex $x_k$ but only a subset of them. We first strengthen the constraint to requiring that $z_{kl}$ form a simple loop and the first configuration $x_1$ must be in the loop. (If a simple open path is desired, a dummy configuration $x_1$ can be introduced by assigning $d_{1k}=d_{k1}=0$.) First, we ensure that there is at most one emanating edge from a vertex:
\begin{equation}
\begin{aligned}
\label{eq:atmostone}
&\sum_{k=1}^Kz_{kl}\leq1\quad\forall l=2,\cdots,K  \\
&\sum_{k=1}^Kz_{k1}=1.
\end{aligned}
\end{equation}
Second, we ensure the number of emanating edges is equal to the number of incident edges:
\begin{align}
\label{eq:sameedge}
\sum_{k=1}^Kz_{kl}=\sum_{k=1}^Kz_{lk}
\quad\forall l=1,\cdots,K.
\end{align}
A major challenge in TSP is to ensure that there are no independent loops (e.g. $z_{23}=z_{34}=z_{42}=1=z_{56}=z_{67}=z_{75}$). To this end, we use the single-commodity flow (SCF) formulation \cite{letchford2013compact}. The idea is to have the robot carry some amount of goods and it needs to unload a unit amount of goods to each selected vertex $\E{x}_k$ (similar to the network-flow constraint). To model this behavior, we introduce continuous variables $g_{kl}$ (we assume $g_{kk}=0$) representing the amount of "goods" the robot is carrying when traveling along $d_{kl}$. If we introduce the following constraint:
\begin{equation}
\begin{aligned}
\label{eq:NO_DAGGLING_LOOP}
&\sum_{k=1}^Kg_{kl}-\sum_{k=1}^Kg_{lk}=\sum_{k=1}^Kz_{kl}\quad\forall l=2,\cdots,K   \\
&0\leq g_{kl}\leq Kz_{kl}\quad \\
&g_{kl}\leq\sum_{a=1}^K\sum_{b=1}^K z_{ab}-1,
\end{aligned}
\end{equation}
then we can guarantee the solution contains only one simple loop. The first line implies that if a vertex $x_k$ is selected, then the robot will carry one unit fewer amount of goods when leaving this vertex. The second line dictates that, if an edge is not selected, then the robot cannot carry any goods along it. The third line requires the robot to carry the minimum amount of goods that is just enough.

Altogether, our MILP formulation solves the following problem:
\begin{equation}
\begin{aligned}
\label{eq: Full MILP formulation - TSP + Irradiation}
\argmin{t_k,z_{kl},g_{kl}}&
\sum_{k=1}^Nt_k + \frac{1}{v_{max}}\sum_{k,l=1}^Kd_{kl}z_{kl} \\
\ST&\mu_i \geq \mu_{min}\quad\forall i=1,\cdots,N \\ 
&\prettyref{eq:bigM},\ref{eq:atmostone},\ref{eq:sameedge},\ref{eq:NO_DAGGLING_LOOP}.
\end{aligned}
\end{equation}
Our implementation uses Gurobi \cite{gurobib}, a fast commercial solver that handles MILPs. Note, however, that for any non-trivial case, this formulation consists of a very large MILP problem, for which finding a solution is often intractable, even with modern powerful solvers. To remedy that, we propose our two-stage approximation scheme that sidesteps the large dimensionality of this problem by dividing it into two parts.
\section{Additional Experiments}

\subsection{2.5D Resolution Experiments}

In 2.5D there are two main factors that could influence the outcomes of our algorithm - the resolution of the vantage point selection grid and the resolution of the environment representation - i.e. - how finely we discretize the polygonal obstacles when calculating their respective irradiances. It is worth noting that finer environment resolutions should yield more accurate estimates of the irradiance at points in the surface, whereas a finer vantage point grid should afford more optimal dwell times. 
In the first experiment, the grid resolution was kept at 0.1m , whereas the environment resolution was changed between $\frac{1}{2} m$ and $\frac{1}{32}m$ (i.e., the line segments of the polygons were subdivided into line segments at the specified environment resolution) and the optimal disinfection trajectories were computed using our algorithm on our 25 randomly generated rooms. In \prettyref{fig:  environment resolution v disinfection 2.5D}, the values were normalized per room by the quantities obtained in the coarsest resolution to facilitate comparisons. We can see that the mean total dwell time did not vary much, while the path lengths grew by 50\% on average, which is expected given the 150\% increase in the number of selected irradiation points. This increase should not greatly impact the total disinfection performance, given the small fraction that the displacement time has on the total disinfection time.

\begin{figure}[htbp] 
\centering
\includegraphics[width=0.8\linewidth]{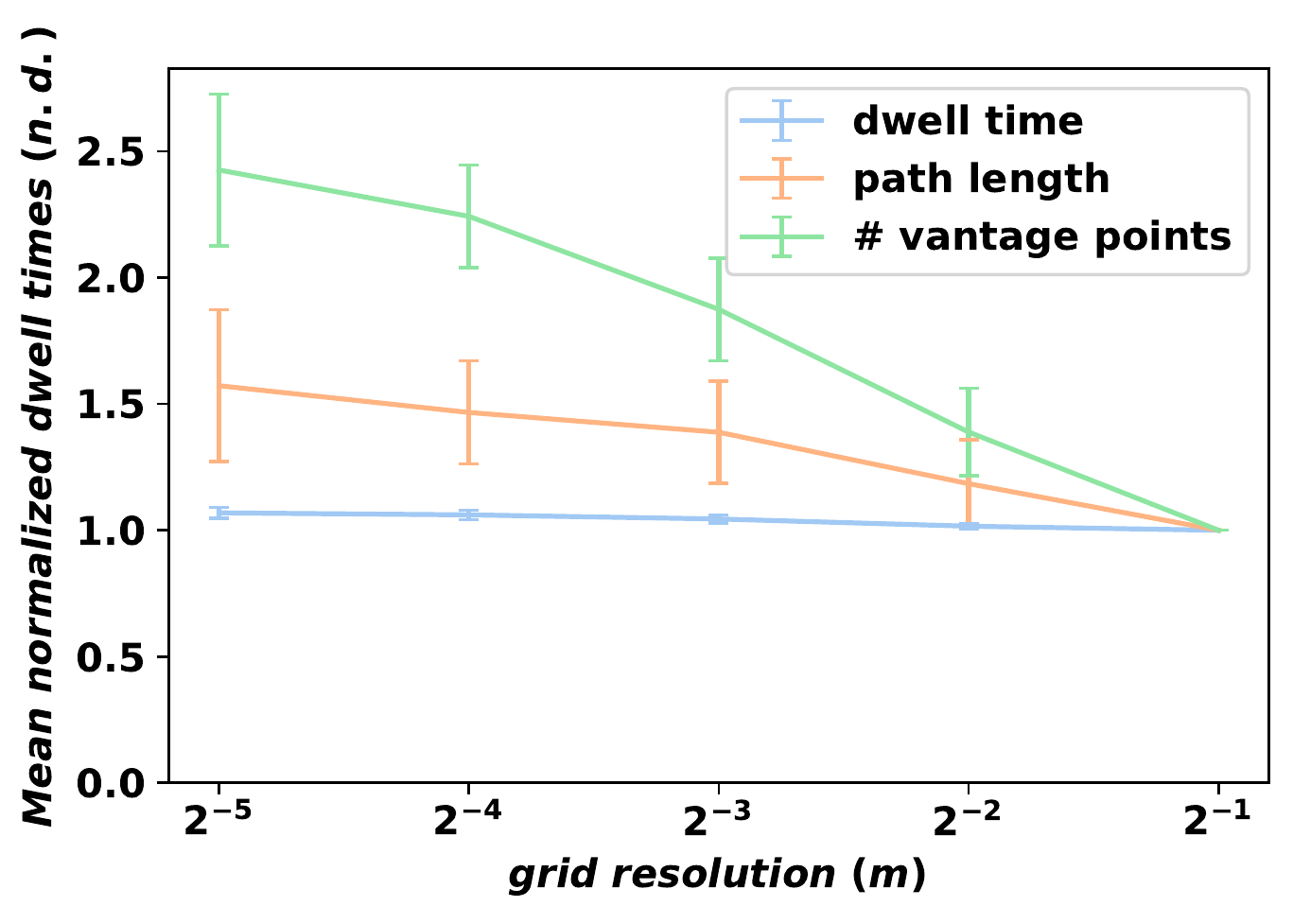}
\caption{Effect of environment resolution on 2.5D disinfection performance metrics - error bars indicate the standard deviation (best seen in color)} \label{fig:  environment resolution v disinfection 2.5D}
\vspace{-12px}
\end{figure}

On the second experiment, we kept the environment resolution fixed at $\frac{1}{8}$ given that it was observed in the previous experiment to be a balanced point between environment description and the time needed to calculate the irradiance matrix. We then varied the vantage point selection grid resolution between $\frac{1}{2} m$ and $\frac{1}{32}m$. The results in \prettyref{fig:  grid resolution v disinfection metrics} were normalized for comparison in a similar fashion to the previous ones. Here, an inverse trend is observed: As the grid resolution gets coarser, the dwell times grow larger, which is expected, as a coarser grid is likely to offer less optimal vantage points. Interestingly, once the resolution reaches $\frac{1}{8} m$ the reduction in dwell times with further refinement is nearly negligible and, similarly to before, finer grids result in a larger number of vantage points being selected, with diminishing returns in terms of total dwell times. 

\begin{figure}[htbp] 
\centering
\includegraphics[width=0.8\linewidth]{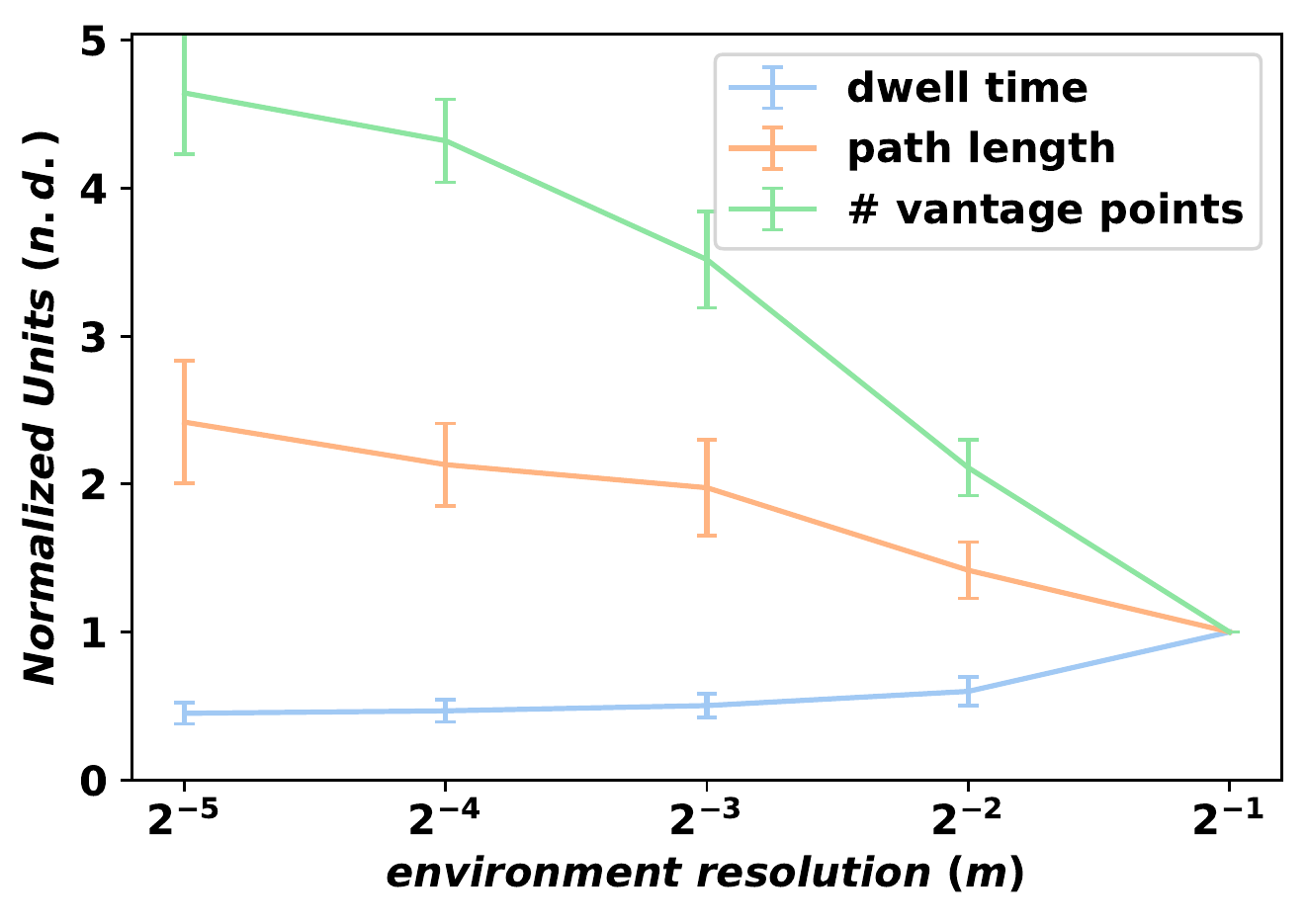}
\caption{Effect of grid resolution on 2.5D disinfection performance metrics - error bars indicate the standard deviation (best seen in color)} \label{fig:  grid resolution v disinfection metrics}
\vspace{-12px}
\end{figure}

\subsection{3D Resolution Experiments}

In order to verify if our conclusions in the simplified 2.5D model held in a 3D environment, we reproduced the environment grid resolution experiments in our 3D CAD model. The 3D grids were created with a spacing of 1000,750, 500, 400, 300 ,250 ,200 ,150 and 125 milimiters and the effects of the grid size on the 30-minute and Asymptotic disinfection tasks were evaluated. Finer resolutions were not tested due to memory limitations on our workstations. \prettyref{fig: Area V res 30 min} shows that, as expected, finer resolutions result in higher surface coverage. However, between 300 and 400 millimeters, strong diminishing returns are observed - in particular when considering the cubic scaling related to the spatial grid in terms of compute time for the irradiance matrix - and in the number of variables in the linear program. One should also note that the 30-minute disinfection coverage for the robots is close to their asymptotic disinfection performance, seen in \prettyref{fig: Area V res assymptotic}. Another curious observation is that it seems that the towerbot design seems to be the least sensitive to the grid resolution. Adding to this tradeoff, we can see in \prettyref{fig: movement V res 30 min} that the end-effector planned end-effector motion grows as the grid spacing is reduced. This, again, is expected, as the number of selected disinfection points in the optimal irradiation points tends to grow with finer environmental resolution - which is also observed in the asymptotic case, seen in \prettyref{fig: movement V res assymptotic}. Figures \ref{fig: floatbot 500mm disinfection} and \ref{fig: floatbot 125 mm disinfection} help illustrate this trade-off. Note how the two trajectories seem to still cover many of the same vantage points - yet \prettyref{fig: floatbot 125 mm disinfection}, illustrating the trajectory for the finer grid discretization has finer movement and better coverage. In sum, our experiments suggest that there exists an ideal tradeoff between grid resolution, computation time and disinfection performance, though experiments in a more varied set of environments are needed to confirm this trend.

\begin{figure}[htbp] 
\centering
\includegraphics[width=0.8\linewidth]{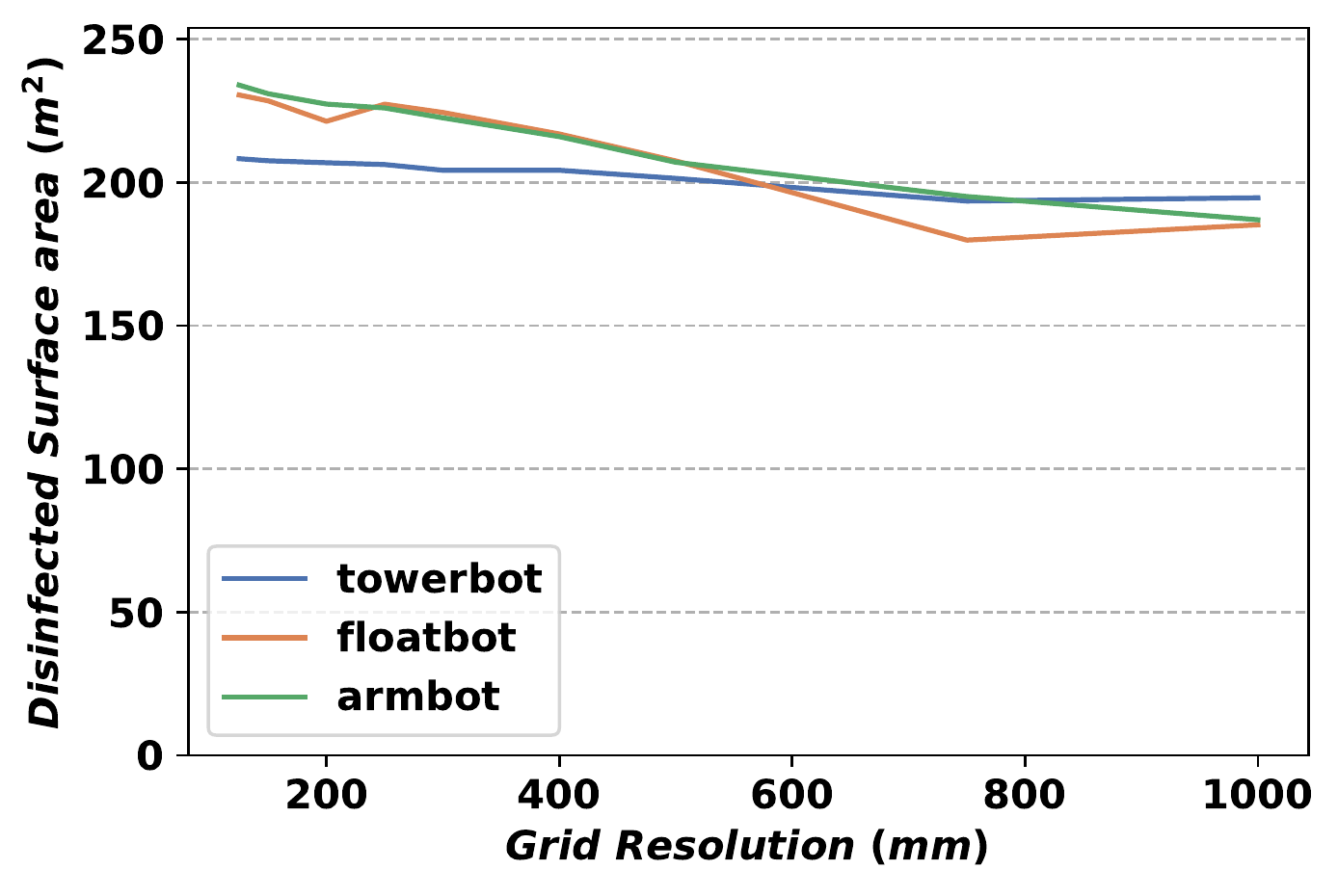}
\caption{Effect of Grid resolution on total disinfected area in 30 minutes budget} \label{fig: Area V res 30 min}
\vspace{-12px}
\end{figure}

\begin{figure}[htbp] 
\centering
\includegraphics[width=0.8\linewidth]{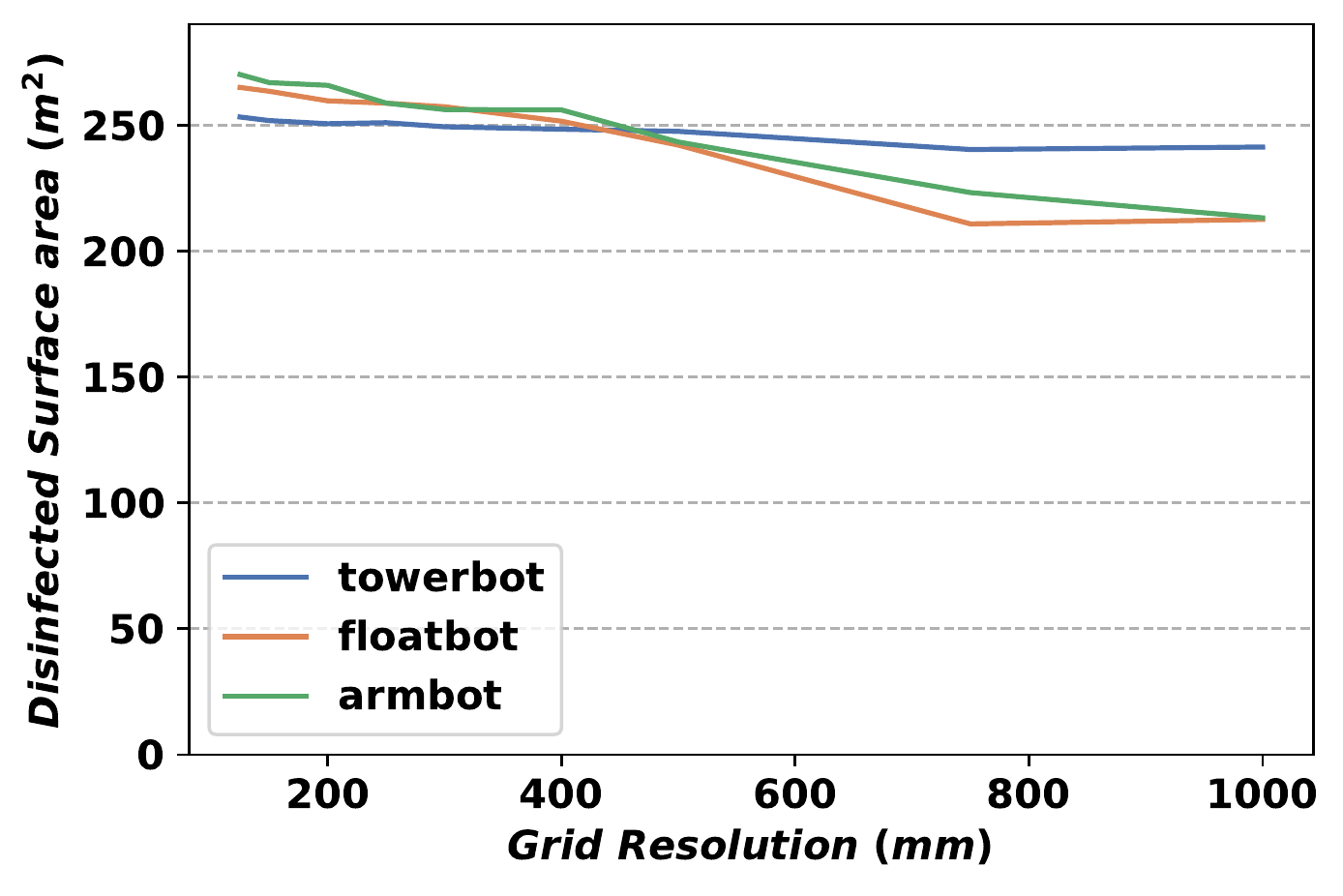}
\caption{Effect of grid resolution on total disinfected area under 100-hours budget (asymptotic)} \label{fig: Area V res assymptotic}
\vspace{-12px}
\end{figure}

\begin{figure}[htbp] 
\centering
\includegraphics[width=0.8\linewidth]{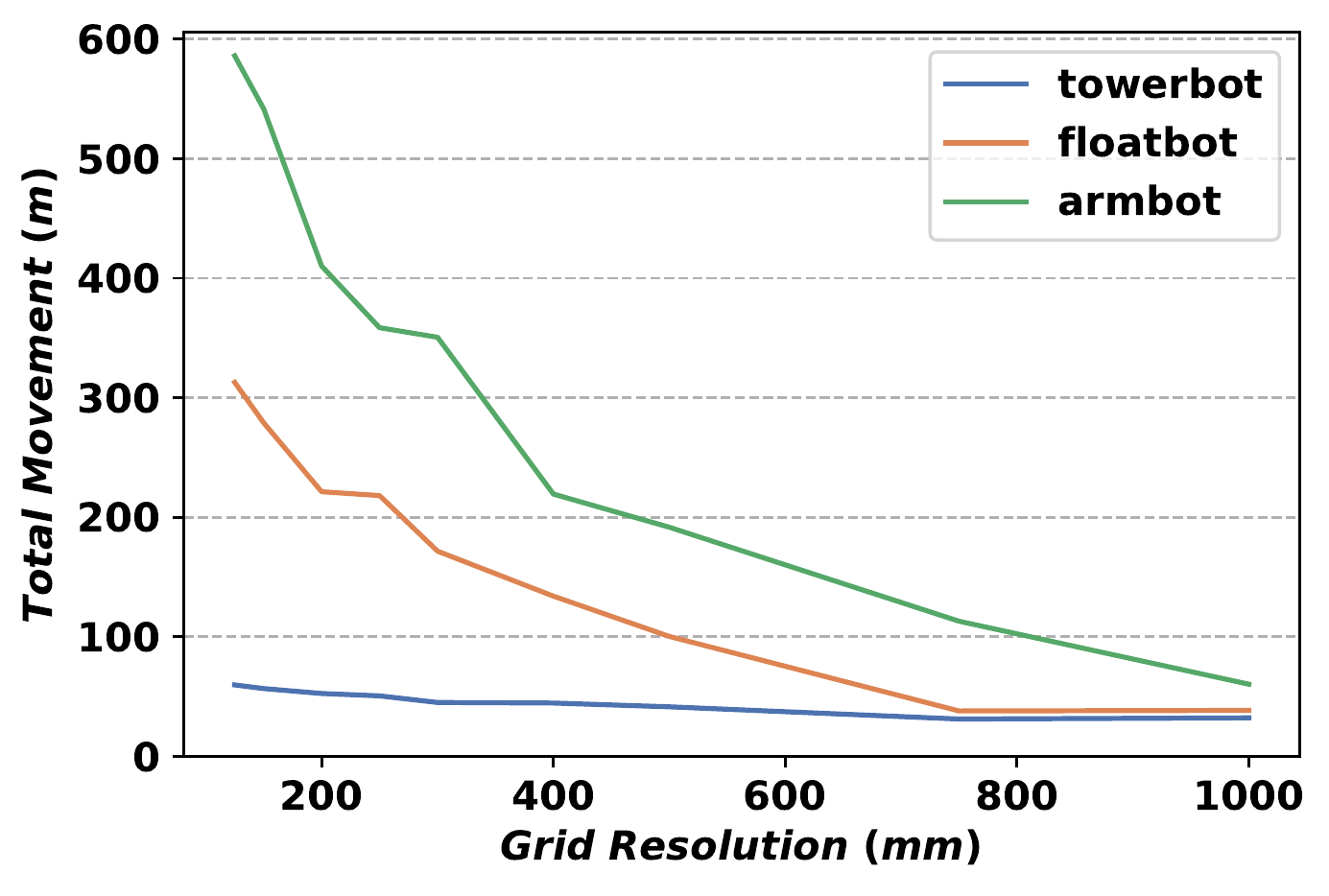}
\caption{Effect of grid resolution on total end-effector movement under 30-minute disinfection budget} \label{fig: movement V res 30 min}
\vspace{-12px}
\end{figure}

\begin{figure}[t!] 
\centering
\includegraphics[width=0.8\linewidth]{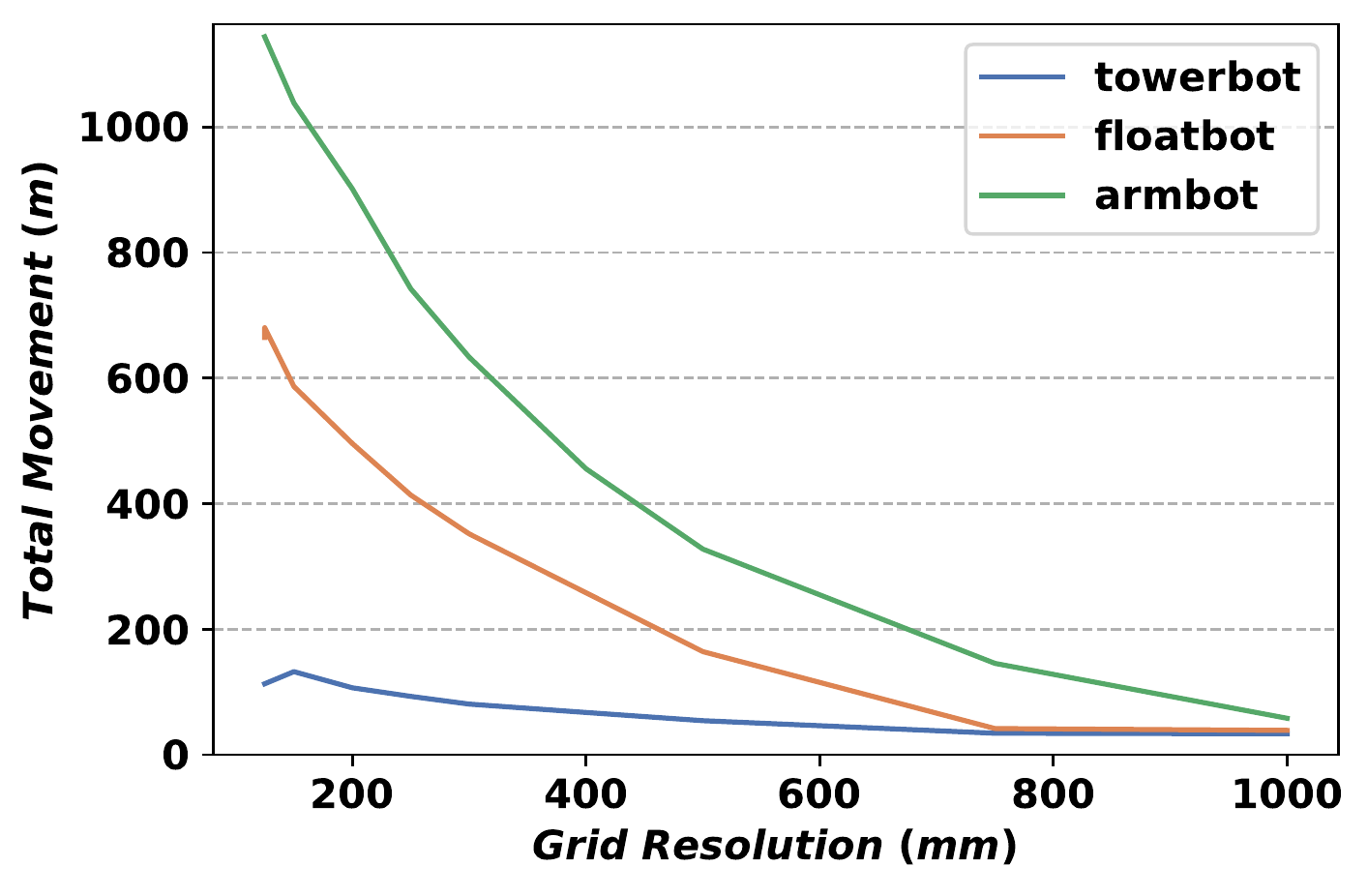}
\caption{Effect of grid resolution on total end-effector movement under 100-hour disinfection budget} \label{fig: movement V res assymptotic}
\vspace{-12px}
\end{figure}

\begin{figure}[htbp] 
\centering
\includegraphics[width=0.9\linewidth]{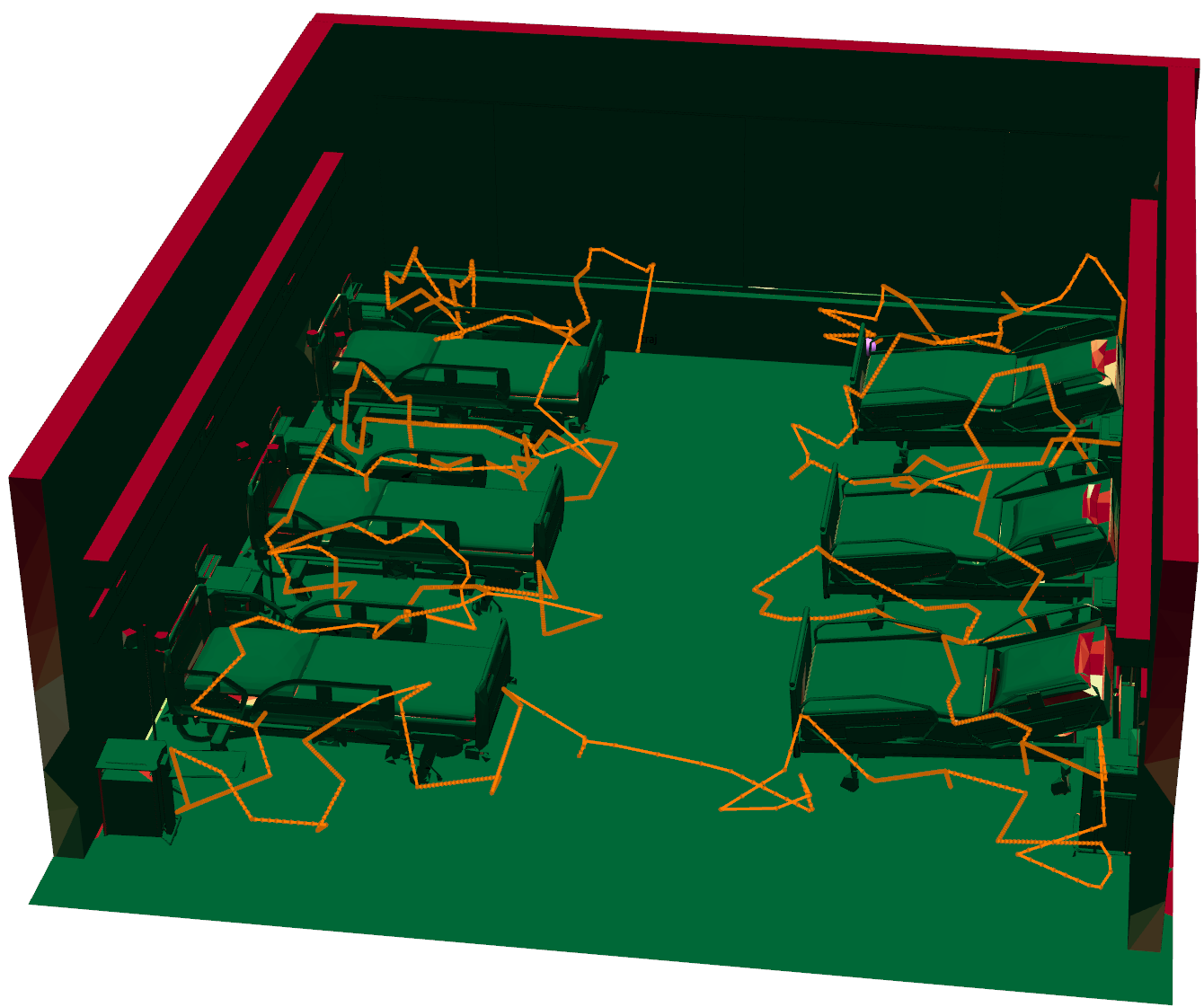}
\caption{Final disinfection tour of Floatbot under 30 minutes time budget with an environment grid resolution of 500mm. Trajectory taken is illustrated in orange. Estimated surface coverage = 82.3\%} \label{fig: floatbot 500mm disinfection}
\vspace{-12px}
\end{figure}

\begin{figure}[t!] 
\centering
\includegraphics[width=0.9\linewidth]{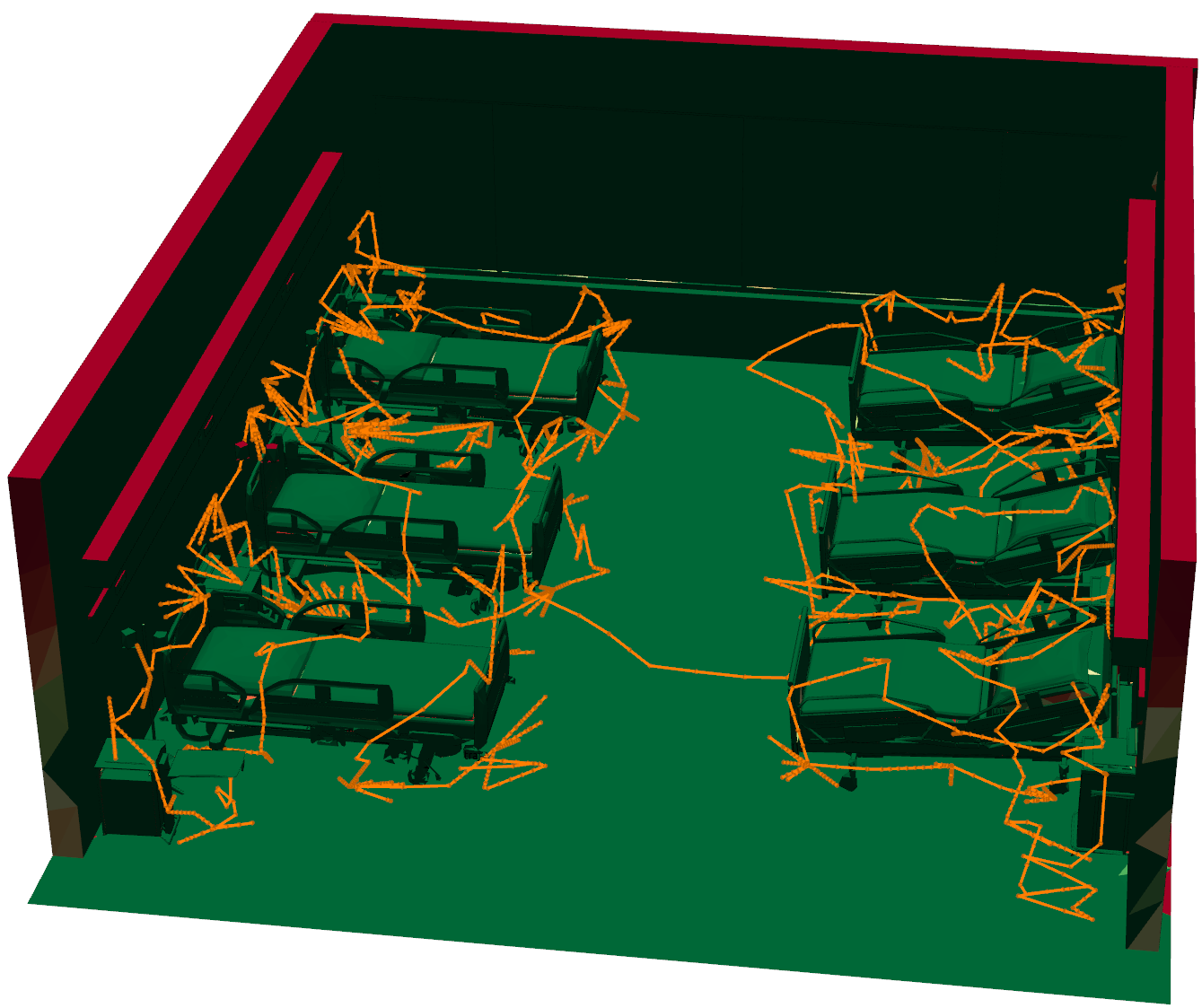}
\caption{Final disinfection tour of Floatbot under 30 minutes time budget with an environment grid resolution of 125mm. Trajectory taken is illustrated in orange. Estimated surface coverage = 90.9\%} \label{fig: floatbot 125 mm disinfection}
\vspace{-12px}
\end{figure}

\fi

\clearpage
\printbibliography

@article{bogaerts2018gradient,
    title = {{A gradient-based inspection path optimization approach}},
    year = {2018},
    journal = {IEEE Robotics and Automation Letters},
    author = {Bogaerts, Boris and Sels, Seppe and Vanlanduit, Steve and Penne, Rudi},
    number = {3},
    pages = {2646--2653},
    volume = {3},
    publisher = {IEEE}
}

@article{romeijn2006new,
    title = {{A new linear programming approach to radiation therapy treatment planning problems}},
    year = {2006},
    journal = {Operations Research},
    author = {Romeijn, H Edwin and Ahuja, Ravindra K and Dempsey, James F and Kumar, Arvind},
    number = {2},
    pages = {201--216},
    volume = {54},
    publisher = {INFORMS}
}

@article{lozano1979algorithm,
    title = {{An algorithm for planning collision-free paths among polyhedral obstacles}},
    year = {1979},
    journal = {Communications of the ACM},
    author = {Lozano-P{\'{e}}rez, Tomás and Wesley, Michael A},
    number = {10},
    pages = {560--570},
    volume = {22},
    publisher = {ACM New York, NY, USA}
}

@article{bircher2017incremental,
    title = {{An incremental sampling-based approach to inspection planning: the rapidly exploring random tree of trees}},
    year = {2017},
    journal = {Robotica},
    author = {Bircher, Andreas and Alexis, Kostas and Schwesinger, Ulrich and Omari, Sammy and Burri, Michael and Siegwart, Roland},
    number = {6},
    pages = {1327--1340},
    volume = {35},
    publisher = {Cambridge University Press}
}

@article{letchford2013compact,
    title = {{Compact formulations of the Steiner traveling salesman problem and related problems}},
    year = {2013},
    journal = {European Journal of Operational Research},
    author = {Letchford, Adam N and Nasiri, Saeideh D and Theis, Dirk Oliver},
    number = {1},
    pages = {83--92},
    volume = {228},
    publisher = {Elsevier}
}

@article{armellino2020comparative,
    title = {{Comparative evaluation of operating room terminal cleaning by two methods: Focused multivector ultraviolet (FMUV) versus manual-chemical disinfection}},
    year = {2020},
    journal = {American Journal of Infection Control},
    author = {Armellino, Donna and Goldstein, Kristine and Thomas, Linti and Walsh, Thomas J and Petraitis, Vidmantas},
    number = {2},
    pages = {147--152},
    volume = {48},
    publisher = {Elsevier}
}

@article{galceran2015coverage,
    title = {{Coverage path planning with real-time replanning and surface reconstruction for inspection of three-dimensional underwater structures using autonomous underwater vehicles}},
    year = {2015},
    journal = {Journal of Field Robotics},
    author = {Galceran, Enric and Campos, Ricard and Palomeras, Narc\'\is and Ribas, David and Carreras, Marc and Ridao, Pere},
    number = {7},
    pages = {952--983},
    volume = {32},
    publisher = {Wiley Online Library}
}

@article{mosher1999eeg,
    title = {{EEG and MEG: forward solutions for inverse methods}},
    year = {1999},
    journal = {IEEE Transactions on Biomedical Engineering},
    author = {Mosher, John C and Leahy, Richard M and Lewis, Paul S},
    number = {3},
    pages = {245--259},
    volume = {46},
    publisher = {IEEE}
}

@article{bedell2016efficacy,
    title = {{Efficacy of an automated multiple emitter whole-room ultraviolet-C disinfection system against coronaviruses MHV and MERS-CoV}},
    year = {2016},
    journal = {Infection Control {\&} Hospital Epidemiology},
    author = {Bedell, Kurt and Buchaklian, Adam H and Perlman, Stanley},
    number = {5},
    pages = {598--599},
    volume = {37},
    publisher = {Cambridge University Press}
}

@article{ezzell1996genetic,
    title = {{Genetic and geometric optimization of three-dimensional radiation therapy treatment planning}},
    year = {1996},
    journal = {Medical Physics},
    author = {Ezzell, Gary A},
    number = {3},
    pages = {293--305},
    volume = {23},
    publisher = {Wiley Online Library}
}

@misc{gurobi,
    title = {{Gurobi Optimizer Reference Manual}},
    year = {2018},
    author = {Gurobi Optimization, L L C},
    url = {http://www.gurobi.com}
}

@misc{gurobib,
    title = {{Gurobi Optimizer Reference Manual}},
    year = {2020},
    author = {Gurobi Optimization, L L C},
    url = {http://www.gurobi.com}
}

@article{lee2003integer,
    title = {{Integer programming applied to intensity-modulated radiation therapy treatment planning}},
    year = {2003},
    journal = {Annals of Operations Research},
    author = {Lee, Eva K and Fox, Tim and Crocker, Ian},
    number = {1-4},
    pages = {165--181},
    volume = {119},
    publisher = {Springer}
}

@article{Helsgaun2000EffectiveHeuristic,
    title = {{Effective implementation of the Lin-Kernighan traveling salesman heuristic}},
    year = {2000},
    journal = {European Journal of Operational Research},
    author = {Helsgaun, Keld},
    number = {1},
    month = {10},
    pages = {106--130},
    volume = {126},
    publisher = {Elsevier Science B.V.},
    doi = {10.1016/S0377-2217(99)00284-2},
    issn = {03772217}
}

@inproceedings{Heng2015EfficientEnvironments,
    title = {{Efficient visual exploration and coverage with a micro aerial vehicle in unknown environments}},
    year = {2015},
    booktitle = {Proceedings - IEEE International Conference on Robotics and Automation},
    author = {Heng, Lionel and Gotovos, Alkis and Krause, Andreas and Pollefeys, Marc},
    number = {June},
    month = {6},
    pages = {1071--1078},
    volume = {2015-June},
    publisher = {Institute of Electrical and Electronics Engineers Inc.},
    doi = {10.1109/ICRA.2015.7139309},
    issn = {10504729}
}

@article{Hijnen2006InactivationReviewb,
    title = {{Inactivation credit of UV radiation for viruses, bacteria and protozoan (oo)cysts in water: A review}},
    year = {2006},
    journal = {Water Research},
    author = {Hijnen, W A M and Beerendonk, E F and Medema, G J},
    number = {1},
    pages = {3--22},
    volume = {40},
    url = {http://www.sciencedirect.com/science/article/pii/S004313540500610X},
    doi = {https://doi.org/10.1016/j.watres.2005.10.030},
    issn = {0043-1354}
}

@inproceedings{Englot2010InspectionStructures,
    title = {{Inspection planning for sensor coverage of 3D marine structures}},
    year = {2010},
    booktitle = {IROS 2010 - Conference Proceedings},
    author = {Englot, Brendan and Hover, Franz},
    pages = {4412--4417},
    isbn = {9781424466757},
    doi = {10.1109/IROS.2010.5648908}
}

@misc{Introduction,
    title = {{Introduction To Mean Shift Algorithm | God, Your Book Is Great !!}},
    url = {https://saravananthirumuruganathan.wordpress.com/2010/04/01/introduction-to-mean-shift-algorithm/}
}

@article{bailey2016opengl,
    title = {{OpenGL Compute Shaders}},
    year = {2016},
    journal = {Oregon State University},
    author = {Bailey, Mike}
}

@article{kavraki1996probabilistic,
    title = {{Probabilistic roadmaps for path planning in high-dimensional configuration spaces}},
    year = {1996},
    journal = {IEEE transactions on Robotics and Automation},
    author = {Kavraki, Lydia E and Svestka, Petr and Latombe, J-C and Overmars, Mark H},
    number = {4},
    pages = {566--580},
    volume = {12},
    publisher = {IEEE}
}

@inproceedings{das_probably_approximately_2011,
    title = {{Probably approximately correct coverage for robots with uncertainty}},
    year = {2011},
    booktitle = {2011 IEEE/RSJ International Conference on Intelligent Robots and Systems},
    author = {Das, C and Becker, A and Bretl, T},
    pages = {1160--1166}
}

@inproceedings{coombe2004radiosity,
    title = {{Radiosity on graphics hardware}},
    year = {2004},
    booktitle = {Proceedings of Graphics Interface 2004},
    author = {Coombe, Greg and Harris, Mark J and Lastra, Anselmo},
    pages = {161--168},
    organization = {Citeseer}
}

@article{heitz2016real,
    title = {{Real-time polygonal-light shading with linearly transformed cosines}},
    year = {2016},
    journal = {ACM Transactions on Graphics (TOG)},
    author = {Heitz, Eric and Dupuy, Jonathan and Hill, Stephen and Neubelt, David},
    number = {4},
    pages = {1--8},
    volume = {35},
    publisher = {ACM New York, NY, USA}
}

@misc{park2007robot,
    title = {{Robot cleaner having floor-disinfecting function}},
    year = {2007},
    author = {Park, Jee-su and Lee, Ju-Sang and Ko, Jang-youn and Cho, Young-ik and Song, Jeong-Gon},
    publisher = {Google Patents}
}

@article{cohen1985hemi,
    title = {{The hemi-cube: A radiosity solution for complex environments}},
    year = {1985},
    journal = {ACM Siggraph Computer Graphics},
    author = {Cohen, Michael F and Greenberg, Donald P},
    number = {3},
    pages = {31--40},
    volume = {19},
    publisher = {ACM New York, NY, USA}
}

@article{bahr1968method,
    title = {{The method of linear programming applied to radiation treatment planning}},
    year = {1968},
    journal = {Radiology},
    author = {Bahr, G K and Kereiakes, J G and Horwitz, H and Finney, R and Galvin, J and Goode, K},
    number = {4},
    pages = {686--693},
    volume = {91},
    publisher = {The Radiological Society of North America}
}

@inproceedings{green2005opengl,
    title = {{The OpenGL framebuffer object extension}},
    year = {2005},
    booktitle = {Game developers conference},
    author = {Green, Simon},
    volume = {2005}
}

@article{4121581,
    title = {{The Solid Angle of a Plane Triangle}},
    year = {1983},
    journal = {IEEE Transactions on Biomedical Engineering},
    author = {Van Oosterom, A and Strackee, J},
    number = {2},
    pages = {125--126},
    volume = {BME-30},
    doi = {10.1109/TBME.1983.325207}
}

@misc{lyon2008uv,
    title = {{Uv Sterilizing Wand}},
    year = {2008},
    author = {Lyon, Donald E},
    publisher = {Google Patents}
}

@article{Hu2020SegmentingEnvironments,
    title = {{Segmenting areas of potential contamination for adaptive robotic disinfection in built environments}},
    year = {2020},
    journal = {Building and Environment},
    author = {Hu, Da and Zhong, Hai and Li, Shuai and Tan, Jindong and He, Qiang},
    month = {10},
    pages = {107226},
    volume = {184},
    publisher = {Elsevier Ltd},
    doi = {10.1016/j.buildenv.2020.107226},
    issn = {03601323},
    pmid = {32868961}
}

@article{Paull2013Sensor-drivenVehicles,
    title = {{Sensor-driven online coverage planning for autonomous underwater vehicles}},
    year = {2013},
    journal = {IEEE/ASME Transactions on Mechatronics},
    author = {Paull, Liam and Saeedi, Sajad and Seto, Mae and Li, Howard},
    number = {6},
    pages = {1827--1838},
    volume = {18},
    publisher = {Institute of Electrical and Electronics Engineers Inc.},
    doi = {10.1109/TMECH.2012.2213607},
    issn = {10834435}
}

@inproceedings{Bircher2015StructuralRobotics,
    title = {{Structural inspection path planning via iterative viewpoint resampling with application to aerial robotics}},
    year = {2015},
    booktitle = {Proceedings - IEEE International Conference on Robotics and Automation},
    author = {Bircher, Andreas and Alexis, Kostas and Burri, Michael and Oettershagen, Philipp and Omari, Sammy and Mantel, Thomas and Siegwart, Roland},
    number = {June},
    month = {6},
    pages = {6423--6430},
    volume = {2015-June},
    publisher = {Institute of Electrical and Electronics Engineers Inc.},
    doi = {10.1109/ICRA.2015.7140101},
    issn = {10504729}
}

@inproceedings{Cheng2008Time-optimalCoverage,
    title = {{Time-optimal UAV trajectory planning for 3D urban structure coverage}},
    year = {2008},
    booktitle = {2008 IEEE/RSJ International Conference on Intelligent Robots and Systems, IROS},
    author = {Cheng, Peng and Keller, James and Kumar, Vijay},
    pages = {2750--2757},
    isbn = {9781424420582},
    doi = {10.1109/IROS.2008.4650988}
}

@article{Heling2020UltravioletStudies.,
    title = {{Ultraviolet irradiation doses for coronavirus inactivation - review and analysis of coronavirus photoinactivation studies.}},
    year = {2020},
    journal = {GMS hygiene and infection control},
    author = {He{\ss}ling, Martin and H{\"{o}}nes, Katharina and Vatter, Petra and Lingenfelder, Christian},
    pages = {Doc08},
    volume = {15},
    publisher = {German Medical Science},
    doi = {10.3205/dgkh000343},
    issn = {2196-5226}
}

@article{Lindblad2019Ultraviolet-CNeeded,
    title = {{Ultraviolet-C decontamination of a hospital room: Amount of UV light needed}},
    year = {2019},
    journal = {Burns},
    author = {Lindblad, Marie and Tano, Eva and Lindahl, Claes and Huss, Fredrik},
    url = {http://www.sciencedirect.com/science/article/pii/S0305417919300920},
    doi = {https://doi.org/10.1016/j.burns.2019.10.004},
    issn = {0305-4179}
}

@article{sillion1994radiosity,
  title={Radiosity \& global illumination},
  author={Sillion, Francois X and Peuch, Claude},
  year={1994},
  publisher={CUMINCAD}
}

@inproceedings{keller1997instant,
  title={Instant radiosity},
  author={Keller, Alexander},
  booktitle={Proceedings of the 24th annual conference on Computer graphics and interactive techniques},
  pages={49--56},
  year={1997}
}

@inproceedings{laine2007incremental,
  title={Incremental Instant Radiosity for Real-Time Indirect Illumination.},
  author={Laine, Samuli and Saransaari, Hannu and Kontkanen, Janne and Lehtinen, Jaakko and Aila, Timo},
  booktitle={Rendering Techniques},
  pages={277--286},
  year={2007}
}

@inproceedings{stefan2013analysis,
  title={Analysis of optimal placement of LED arrays for visible light communication},
  author={Stefan, Irina and Haas, Harald},
  booktitle={2013 IEEE 77th Vehicular Technology Conference (VTC Spring)},
  pages={1--5},
  year={2013},
  organization={IEEE}
}

@article{zhang2013lighting,
  title={Lighting design for globally illuminated volume rendering},
  author={Zhang, Yubo and Ma, Kwan-Liu},
  journal={IEEE transactions on visualization and computer graphics},
  volume={19},
  number={12},
  pages={2946--2955},
  year={2013},
  publisher={IEEE}
}

@article{extended_report,
  title  = "A Targeted Approach to UV Disinfection of Surfaces",
  author = "Jo\~ao Marcos Correia Marques, Ramya Ramalingam, Zherong Pan and Kris Hauser",
  journal= "https://uofi.box.com/s/gk7o1vyqkiggayquwkeedkhpk05j78oi",
  url    = "https://uofi.box.com/s/gk7o1vyqkiggayquwkeedkhpk05j78oi",
  year   = "2020"
}

\end{document}